\pdfoutput=1

\documentclass[11pt]{article}

\usepackage[final]{acl}

\usepackage{times}
\usepackage{latexsym}
\usepackage{url}
\usepackage{breakurl}
\usepackage{amsmath}
\usepackage{amsfonts}
\usepackage{multirow}
\definecolor{CBlue}{RGB}{9, 60, 146}
\usepackage{booktabs,multirow}
\usepackage[table]{xcolor}
\definecolor{lightrow}{gray}{0.93}

\definecolor{myLightBlue}{HTML}{E6F2FF}
\definecolor{myLightGreen}{HTML}{E8F5E9}
\definecolor{paleGrayGreen}{HTML}{E4EDE4}
\usepackage[T1]{fontenc}
\usepackage{subcaption}
\usepackage{footnote}

\usepackage[utf8]{inputenc}
\usepackage{caption} 

\usepackage{microtype}

\usepackage{inconsolata}

\usepackage{graphicx}
\usepackage{booktabs}
\usepackage[dvipsnames]{xcolor}

\title{Minority-Aware Satisfaction Estimation in Dialogue Systems via Preference-Adaptive Reinforcement Learning}

\author{Yahui Fu, Zi Haur Pang, Tatsuya Kawahara \\ Graduate School of Informatics, Kyoto University, Japan\\ \texttt{\{fu,pang,kawahara\}@sap.ist.i.kyoto-u.ac.jp}}

\begin{document}
\maketitle

\begin{abstract}
User satisfaction in dialogue systems is inherently subjective. When the same response strategy is applied across users, minority users may assign different satisfaction ratings than majority users due to variations in individual intents and preferences. However, existing alignment methods typically train one-size-fits-all models that aim for broad consensus, often overlooking minority perspectives and user-specific adaptation. We propose a unified framework that models both individual- and group-level preferences for user satisfaction estimation. First, we introduce Chain-of-Personalized-Reasoning (\textbf{CoPeR}) to capture individual preferences through interpretable reasoning chains. Second, we propose an expectation--maximization--based Majority-Minority Preference-Aware Clustering (\textbf{M\textsuperscript{2}PC}) algorithm that discovers distinct user groups in an unsupervised manner to learn group-level preferences. Finally, we integrate these components into a preference-adaptive reinforcement learning framework (\textbf{PAda-PPO}) that jointly optimizes alignment with both individual and group preferences. 
Experiments on the Emotional Support Conversation dataset demonstrate consistent improvements in user satisfaction estimation, particularly for underrepresented user groups.\footnote{Our source code is publicly available at: \url{https://github.com/fuyahuii/minority-aware-se}.}


\end{abstract}

\section{Introduction}
Personalized dialogue systems that adapt to individual user preferences are crucial for enhancing user satisfaction in human–AI interactions. Accordingly, accurately evaluating whether a dialogue system meets diverse user needs is essential.
\begin{figure}[t]
    \centering
\includegraphics[width=\linewidth]{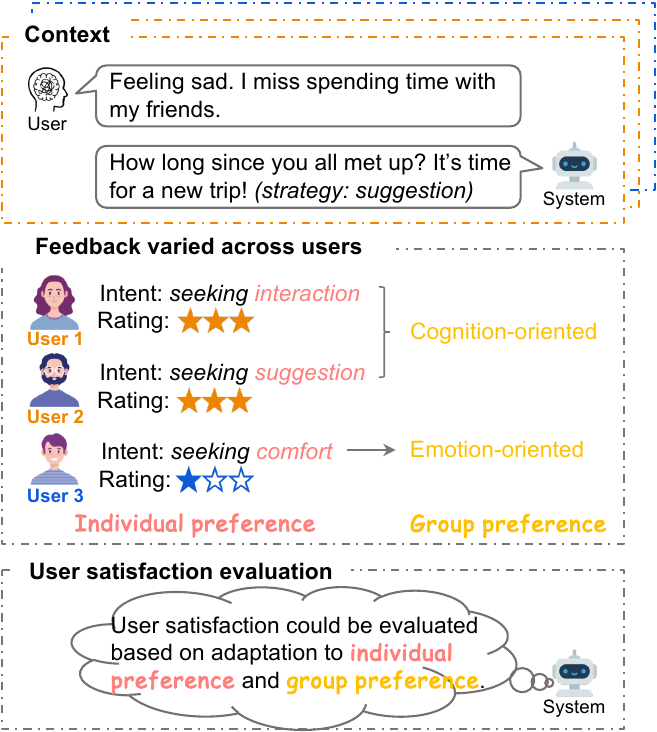}
    \caption{
    \textcolor{orange}{Majority} and \textcolor{CBlue}{minority} users may assign different satisfaction ratings to system responses employing the same strategy due to varying individual intents and preferences. 
    Additionally, users within the same group may exhibit similar preference patterns (e.g., cognition-oriented versus emotion-oriented strategies). This suggests that modeling both individual-specific and group-level preferences could be an effective approach for evaluating user feedback in dialogue systems.
    }
    \label{fig:main}
\end{figure}
Previous research has primarily evaluated dialogue systems using output-centric criteria such as informativeness, relevance, and empathy \cite{bert-score,fu2023dual,xu2024multi}, as well as user-centric metrics such as satisfaction estimation \cite{choi2019convSAT,see2021understanding,lin-etal-2024-interpretable}. However, satisfaction is inherently subjective: as illustrated in Figure~\ref{fig:main}, users may assign divergent satisfaction ratings to the system response employing the same strategy, depending on their individual intents and preferences. 

Reinforcement learning (RL) with a reward model has become a key approach for aligning language models with human preferences. However, existing reward models typically rely on aggregated human judgments \cite{touvron2023llama} (e.g., majority voting or averaging), resulting in universal reward functions that overlook minority preferences and lack personalization. This induces preference collapse, where outputs maximize majority preferences while suppressing minority views \cite{xiao2024algorithmic, yang2024llm, slocum2025diverse}. 
Recent methods tried to solve it by considering the diversity of human preferences into reward modeling \cite{wang2023aligning, jang2023personalized,chakraborty2024maxmin}, but primarily aim to train a one-size-fits-all system that is less controversial. In contrast, we focus on user adaptation by modeling both individual-specific preferences and group-level trends to estimate satisfaction across both minority and majority populations. 

Specifically, we introduce a User-specific Chain-of-Thought (\textbf{UCoT}) prompting strategy and synthesize Chain-of-Personalized-Reasoning (\textbf{CoPeR}) outputs, which capture individual user preferences through explicit reasoning, linking the seeker's intent, the supporter's strategy, and the resulting satisfaction. This enables the supervised fine-tuning (SFT) model to acquire interpretable and user-aligned reasoning capabilities. Since users' majority or minority status is unknown in real-world scenarios, predefined group supervision is impractical. To address this, we propose a Majority-Minority Preference-Aware Clustering (\textbf{M\textsuperscript{2}PC}) module, built upon the Expectation-Maximization (EM) algorithm \cite{dempster1977maximum}, which routes users into majority or minority groups in an unsupervised manner by comparing model perplexities over their dialogues. Separate models are then trained to capture group-specific preferences. Finally, we integrate the individual-level SFT model and the group-specific models as the policy and reference models, respectively, within a Proximal Preference Optimization (PPO) \cite{schulman2017proximal} framework. This results in our Preference-Adaptive Reinforcement Learning (\textbf{PAda-PPO}) approach, which jointly optimizes for both individual and group-level preferences in user satisfaction estimation. 
Our main contributions are:
\begin{itemize}
    \item To the best of our knowledge, we present the first framework that models both individual- and group-level preferences for satisfaction estimation, capturing user diversity by distinguishing majority and minority preferences. 
    \item We propose \textbf{UCoT} prompting and \textbf{CoPeR} synthesis, two CoT-based methods that infer individual satisfaction through explicit reasoning over user intent and system strategy. 
    \item We introduce \textbf{M\textsuperscript{2}PC}, an EM-based unsupervised module that clusters users into majority and minority groups by dialogue perplexity, capturing group-level preference trends.
    \item We develop \textbf{PAda-PPO}, a preference-adaptive reinforcement learning framework that integrates individual- and group-specific models to optimize satisfaction while preserving diverse user preferences.
\end{itemize}

\section{Related Work}
\subsection{User Satisfaction Estimation}

Accurately predicting user satisfaction is essential for evaluating and improving conversational systems. Prior research has focused on perspectives such as sentiment analysis \cite{song-etal-2019-using}, context- and dynamics-aware modeling \cite{choi2019convSAT, see2021understanding, deng-etal-2022-user,ye2023modeling}, and LLM-based frameworks that improve interpretability by inducing human-readable rubrics or dialogue-strategy features \citep{lin-etal-2024-interpretable,kim-etal-2025-llm}.
However, existing approaches often model all users jointly, which tends to suppress minority preferences. 

\subsection{Chain-of-Thought Prompting}
Chain-of-Thought (CoT) prompting has been widely used with LLMs to improve task performance by encouraging step-by-step reasoning \cite{wei2022chain, chae2023dialogue, luong2024reft, xie2025leveraging}. For instance, 
\citet{zhang2024escot} introduce ESCoT, which augments emotional support dialogues with emotion- and strategy-focused CoTs to emulate human emotional reasoning. Differently, we leverage CoT prompting to elicit interpretable reasoning traces that explain how user preferences influence the satisfaction score.


\begin{figure*}
    \centering
\includegraphics[width=\textwidth]{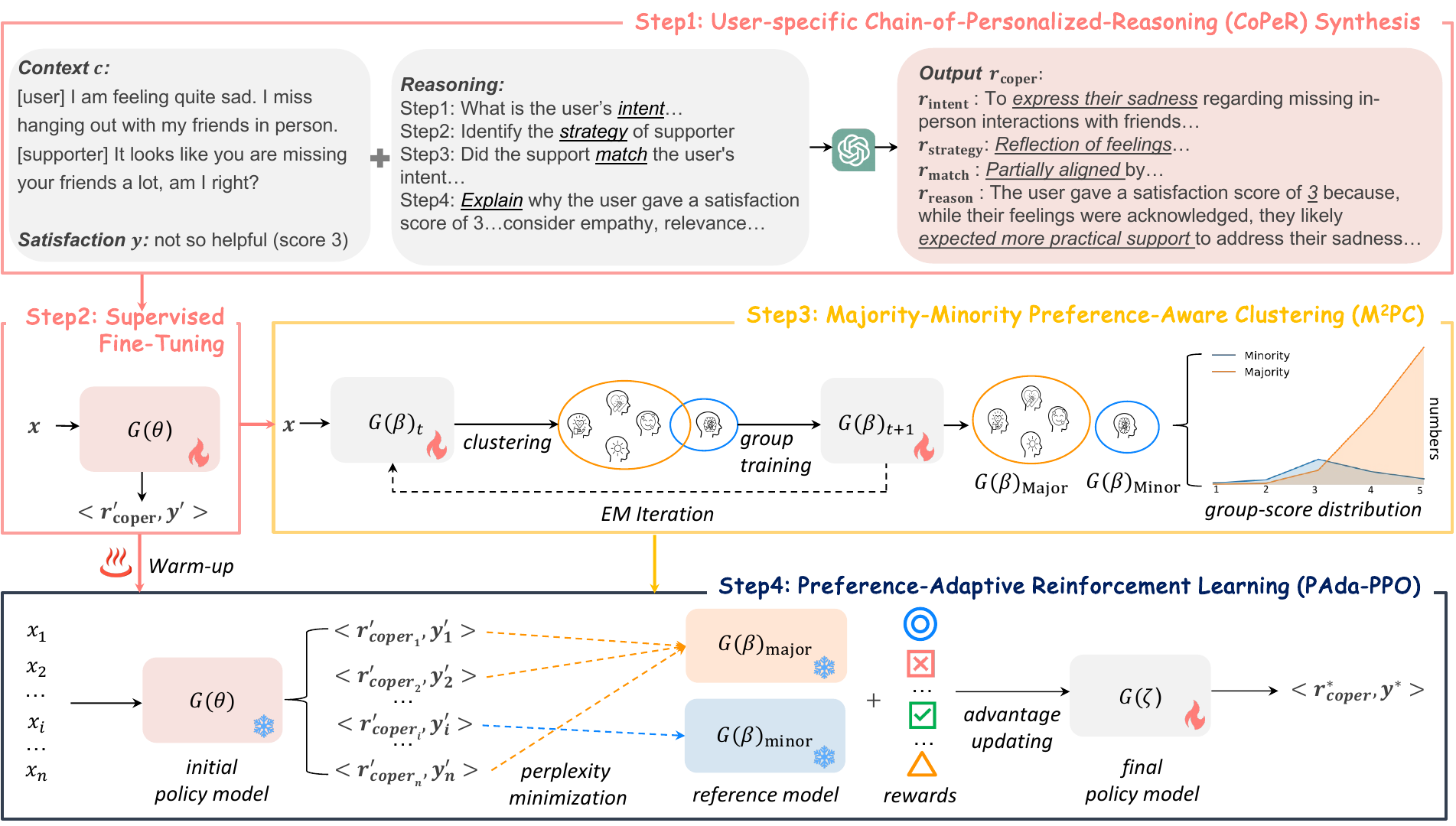}
    \caption{The architecture of our proposed method for the user's satisfaction $y$ estimation, $x$ is the concatenation of context $c$ and \hyperref[sec:UCOT_prompts]{UCoT prompt} $r_\text{ucot}$.}
    \label{fig:arch}
\end{figure*}

\subsection{Aligning Language Models with Diverse Human Preferences}

Recent work has incorporated preference diversity into reward modeling for RL-based alignment \cite{jang2023personalized, wang2023aligning, li2024personalized, chakraborty2024maxmin, wang2024arithmetic, yang2024rewards, la2025fairness}. For example, \citet{wang2023aligning} model rewards as a posterior over annotator opinions to capture disagreement, while \citet{chakraborty2024maxmin} optimize for minority preferences via MaxMin-RLHF based on the Egalitarian principle. Other approaches, such as Personalized-RLHF \cite{li2024personalized} and Personalized Soups \cite{jang2023personalized}, adapt to diverse users by attaching user embeddings or merging specialist policy models. 

However, these methods either aim for a one-size-fits-all solution or maintain multiple policies for different groups. In contrast, we train a single policy model that adapts to users by jointly modeling individual and group-level preferences. In addition, prior work emphasizes reward modeling, RL training also depends on KL regularization with respect to a reference model, typically initialized from the SFT model. As the SFT model tends to overfit to majority behaviors, the reference model may inherit this bias, limiting its ability to guide the policy across diverse users. 

We address this limitation from the source by enhancing preference awareness at both SFT and RL stages. We first model individual preferences via UCoT/ CoPeR during SFT, and then derive separate majority/minority reference models through M\textsuperscript{2}PC clustering, enabling KL regularization to better reflect user diversity during RL.

\section{Preliminary}
We conducted our experiments on the Emotional Support Conversation (ESConv) dataset \cite{liu2021towards}. Throughout each conversation, help-seekers (users) provided feedback every two utterances received from the supporter, rating the helpfulness of these messages on a five-star scale. We denote this feedback score as ``satisfaction score.'' We categorized feedback scores of 3 or lower as ``low satisfaction'' and scores higher than 3 as ``high satisfaction.'' There is no clear, universally accepted rule for partitioning users, factors such as race, age, gender, and personality will create highly varied preferences \cite{aroyo2023dices, chakraborty2024maxmin,fu2024styemp}. Therefore, we heuristically classified users whose proportion of high satisfaction scores exceeded 60\% as belonging to a ``majority'' population, with the remainder designated as the ``minority.'' Consequently, 81.4\% of conversations were identified as majority, while 18.6\% fell into the minority category. 

\section{Proposed Method}
In this work, we tackle user satisfaction prediction in dialogue, given a user's input and a supporter's response. As shown in Figure~\ref{fig:arch}, our framework consists of four main steps.

We first introduce UCoT prompt and use GPT-4.1-mini\footnote{\href{https://platform.openai.com/docs/models/gpt-4.1-mini}{https://platform.openai.com/docs/models/gpt-4.1-mini}} for user-specific CoPeR synthesis, then fine-tune a base model $G(\theta)$ on this reasoning to equip it with user-specific inference ability. Subsequently, we introduce M\textsuperscript{2}PC algorithm to group users into two distinct clusters, training a separate model for each. Finally, these group-specific models serve as reference models in the proposed PAda-PPO framework.

\subsection{User-specific CoPeR Synthesis}
\subsubsection{UCoT Prompt}
\label{sec:UCOT_prompts}


Drawing on findings that user satisfaction hinges on correctly identifying the user’s intent and deploying responses that appropriately match the intent and evolving needs \cite{deng-etal-2022-user, fu2023reasoning, lin-etal-2024-interpretable}, we design a User-specific Chain-of-Thought (\textbf{UCoT}) prompt to make this reasoning explicit for satisfaction estimation. Given a dialogue context, the model is guided to: (1) infer the user's underlying intent; (2) identify the supporter's primary response strategy (e.g., \textit{Question}, \textit{Reflection of Feelings}); (3) evaluate whether the strategy aligns with the user's need; and (4) predict the user's feedback score by considering factors such as empathy and relevance. Details of the prompt appear in Appendix~\ref{sec:prompts_example} (Figure~\ref{fig:prompt_cot}).

\subsubsection{CoPeR Synthesis}

To train a model with the reasoning ability from steps (1)–(3) to correctly predict the user’s feedback score, we use GPT-4.1-mini to synthesize reasoning rationales conditioned on the user’s actual feedback score. Consequently, step (4) is to “explain the rationale behind the user’s feedback score in terms of emotional and practical relevance.” Further details are provided in Appendix \ref{sec:prompts_example} (Figure \ref{fig:prompt_coper}). This structured format enables more interpretable synthesis of user-centered reasoning behind feedback scores.

\subsection{Supervised Fine-Tuning}
After obtaining the synthesized CoPeR output $r_{\text{coper}}$, which comprises four components: $r_{\text{intent}}$, $r_{\text{strategy}}$, $r_{\text{match}}$, and $r_{\text{reason}}$ (as illustrated in Figure~\ref{fig:arch}), along with the UCoT prompt $r_\text{ucot}$, we fine-tune a base model on a dataset of input-target pairs formatted as $(c+r_\text{ucot}, r_{\text{coper}}+y)$. Formally, the output generation process can be decomposed into a sequence of next token prediction actions, denoting as $e = [a_1, a_2, \dots, a_T, s, \text{<eos>}]$. The training objective is defined as:
\begin{equation}
\begin{aligned}
\mathcal{L}_{\text{SFT}}(\theta) 
= & - \sum_{t=1}^{T} \log G_\theta(a_t \mid c+r_\text{ucot}, a_{<t}) \\
  & - \log G_\theta(s \mid c+r_\text{ucot}, a_{1:T}),
\end{aligned}
\end{equation}
where $a_t$ is the token in the vocabulary, $s$ represents the satisfaction score, and $\text{<eos>}$ indicates the end-of-sequence token.
The first term supervises the generation of user-specific reasoning ($r_{\text{coper}}$), and the second term guides the model to accurately predict the satisfaction score ($s$) conditioned on the generated reasoning, and SFT stands for supervised fine-tuning. This warm-up phase equips the model with interpretable and user-aligned reasoning capabilities to generate a proper response. 

\subsection{M\textsuperscript{2}PC}
\label{sec:em}
Human preferences in dialogue vary due to factors such as socio-cultural background~\cite{aroyo2023dices,chakraborty2024maxmin}, personality~\cite{richendoller1994exploring}, and gender~\cite{costa2014associations}. To model such diversity, we first examine preference differences between majority and minority users according to the supporter strategies, then apply an unsupervised EM-based clustering method to separate them.

\subsubsection{Preference Divergence Across Groups}
\label{subsec:user_pref}
\begin{figure}[t]
  \centering
  \begin{subfigure}{\linewidth}
    \centering
    \includegraphics[width=0.9\linewidth]{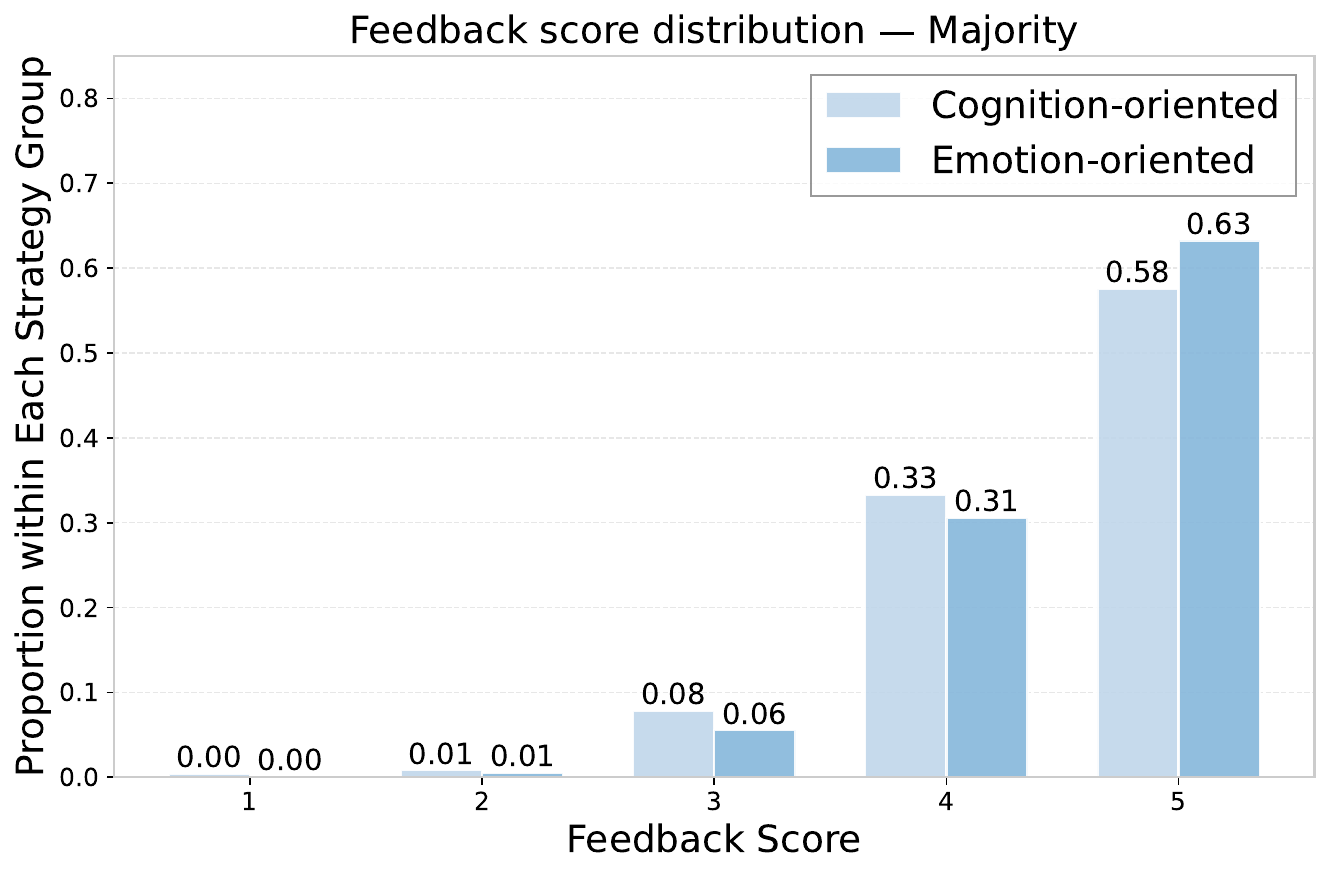}
    \label{fig:feedback_majority}
  \end{subfigure}
  \vspace{0.8em}
  \begin{subfigure}{\linewidth}
    \centering
    \includegraphics[width=0.9\linewidth]{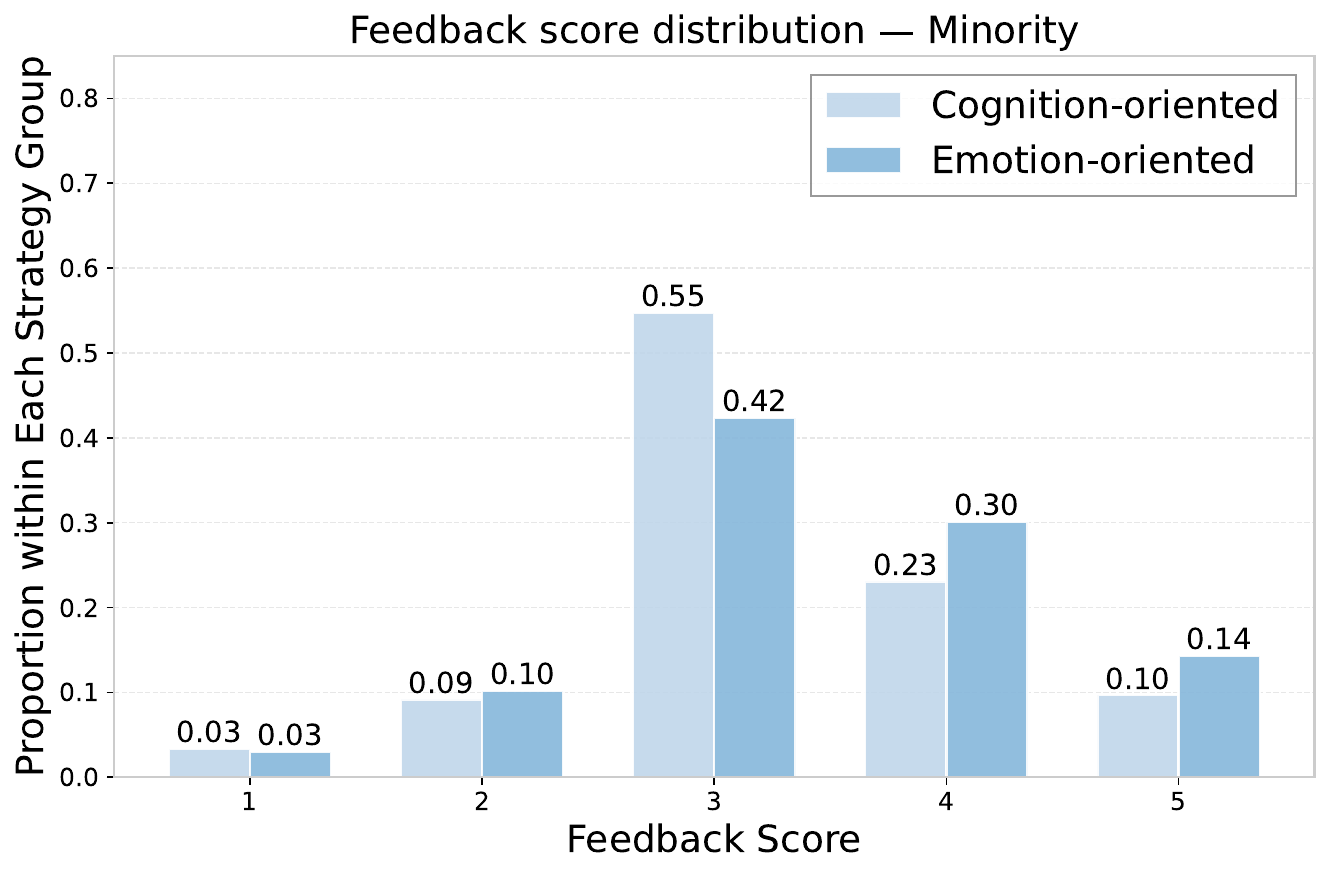}
    \label{fig:feedback_minority}
  \end{subfigure}
  \caption{Distribution of feedback scores for cognition‑ and emotion‑oriented supporter response strategies across majority and minority user groups from ESConv dataset.}
  \label{fig:feedback_compare}
\end{figure}

We categorize supporter strategies in the ESConv dataset into two strategy types: Cognition-oriented, including \textit{Question}, \textit{Restatement or Paraphrasing}, \textit{Providing Suggestions}, and \textit{Information}; and Emotion-oriented, including \textit{Reflection of Feelings}, \textit{Self-disclosure}, and \textit{Affirmation and Reassurance}. We then computed the score distributions (1–5) for each strategy within the majority and minority user groups (Figure~\ref{fig:feedback_compare}).

Across both majority and minority groups, emotion-oriented strategies are more likely to receive high feedback (0.94 > 0.91; 0.44 > 0.33), with the effect especially pronounced in the minority group. This preference divergence underscores the need for personalized modeling that accounts for group-specific preferences.\footnote{Aggregated proportions reported in the text are obtained by summing the bars for scores 4 and 5 (e.g., 0.94 = 0.63 + 0.31 for emotion-oriented strategies in the majority group), further details are provided in Appendix~\ref{sec:feedback_dis}.}



\subsubsection{Diversity-Aware Clustering}
We adopt an EM strategy that divides users into majority and minority based on the perplexity of their dialogue data under different models.
Each model is further fine-tuned on dialogues from the belonging cluster, separately.
Formally, the algorithm proceeds as follows:

\noindent\textbf{E-step (Expectation):} Given two cluster models at iteration $t$, denoted by $G(\beta)_\text{Major}^{(t)}$ and $G(\beta)_\text{Minor}^{(t)}$ (both initialized from the base model $G(\theta)$ trained from the SFT stage), each user $i$ with dialogue set $D_i$ is assigned to the model with lower perplexity:
\begin{equation}
l_i^{(t)} = \arg\min_{k \in \{\text{Major}, \text{Minor}\}} \text{PPL}(D_i; G(\beta)_k^{(t)}),
\end{equation}
where $l_i^{(t)}$ denotes the cluster assignment of user $i$ at the $t$-th EM iteration, and the perplexity is:
\begin{equation}
\text{PPL}(D_i;\!G(\beta))\!=\!\exp\!\left(\!-\frac{1}{|D_i|}\!\sum_{w\in D_i}\!\log\!P_{G(\beta)}\!(w)\!\right)
\end{equation}
This step is to route users toward the model most closely aligned with their group preferences.

\noindent\textbf{M-step (Maximization):} After cluster assignments are updated, we fine-tune each model $G(\beta)_k$ on dialogues assigned to its respective cluster:
\begin{equation}
G(\beta)_k^{(t+1)} = \arg\min_{G(\beta)_k^{(t)}} \sum_{i: l_i^{(t)} = k} \mathcal{L}(G(\beta)_k^{(t)}; D_i),
\end{equation}
where $\mathcal{L}$ is the negative log-likelihood loss over the dialogue set $D_i$.
This EM process iteratively refines cluster assignments and model parameters, enabling the system to unsupervised discover latent user groups that reflect diverse preference patterns.



\subsection{PAda-PPO}
Leveraging M\textsuperscript{2}PC, each input within a batch is routed to either $G(\beta)_\text{Major}$ or $G(\beta)_\text{Minor}$ based on the model yielding lower perplexity, as shown in Figure~\ref{fig:arch}. We employ PPO \citep{schulman2017proximal} with a clipped objective algorithm for training. Following \citet{luong2024reft}, the value model $V_{\phi}$ is contructed by appending a linear value head on top of the last hidden states of the policy model $G(\xi)$, which is initialized from the SFT model $G(\theta)$. 

\paragraph{Reward Modeling} Given a dataset consisting of \text{(input, target)} pairs $(c +r_\text{ucot}, r_{\text{coper}}+y)$, and generated output as $e = [a_1, a_2, \dots, a_T, s, \texttt{<eos>}]$, the reward function is as follows:
\begin{equation}
r_T = \begin{cases}
+1 & s=y,\\
-1 & \text{otherwise}.
\end{cases}
\end{equation}
The reward of 0 is given for the intermediate tokens ($r_t = 0$ for $t<=T$), such partial reward can help reduce the effect of learning  from sparse reward \cite{trott2019keeping}. 

\paragraph{Diversity-aware KL Regularization}  A divergence penalty was utilized to prevent the policy from diverging significantly from human-like reference behaviors in each group. At each timestep $t$, the Kullback-Leibler (KL) divergence \cite{kullback1951information} between the current RL policy $G(\xi)$ and the corresponding reference policy $G(\beta)_m$ is computed as:
\begin{equation}
\mathrm{KL}_t = D_{\mathrm{KL}}\bigl(G(\xi)(\cdot \mid s_t, m) \parallel G(\beta)_m(\cdot \mid s_t)\bigr),
\label{eq:kl}
\end{equation}
where $G(\beta)_m$ corresponds to $G(\beta)_\text{Major}$ or $G(\beta)_\text{Minor}$ according to the perplexity routing, $m \in \{\text{Major}, \text{Minor}\}$. $s_t$ comprises of all tokens in the \text{input} and all tokens generated so far. In addition, following \citet{zheng2023secrets,luong2024reft}, the total reward at each timestep combines the reward function score and KL penalty terms:
\begin{equation}
r_t^{\text{total}} = r_t - \lambda_{\text{KL}} \cdot \mathrm{KL}_t,
\end{equation}
where $\lambda_{\text{KL}}$ controls penalty strength.

\paragraph{Optimization Objective} 
Following Generalized Advantage Estimation \cite{schulman2015high}, we compute advantages $\hat{A}_t$:
\begin{equation}
\begin{aligned}
\delta_t &= r_t^{\text{total}} + \gamma V_{\phi_{\text{old}}}(s_{t+1}) - V_{\phi_{\text{old}}}(s_t),\\
\hat{A}_t &= \sum_{l=0}^{T-t}(\gamma\lambda)^l \delta_{t+l},
\end{aligned}
\end{equation}
with discount factor $\gamma \in(0,1]$ and $\lambda \in(0,1]$. The value function $V_{\phi}(s_t)$ is estimated by a value head jointly trained with the policy. $\mathcal{L}_{\text{value}}(\phi)$ minimizes deviation between value estimate $V_{\phi}(s_t)$ and return estimate $\hat{R}_t = \hat{A}_t + V_{\phi_{\text{old}}}(s_t)$:
\begin{equation}
\begin{aligned}
&\mathcal{L}_{\text{value}}(\phi)
= \frac{1}{2}\,\mathbb{E}_{t}\!\Bigl[
\max\!\Bigl(
(V_{\phi}(s_t)\!-\!\hat{R}_t)^2,\;\\
&\bigl(V_{\phi_{\text{old}}}(s_t)\!+\!
\operatorname{clip}\!\bigl(
V_{\phi}(s_t)\!-\!V_{\phi_{\text{old}}}(s_t),\!-\epsilon,\!\epsilon
\bigr)\!-\!\hat{R}_t\bigr)^2
\Bigr)\!\Bigr].
\end{aligned}
\end{equation}
The final policy objective is:
\begin{equation}
\begin{aligned}
\mathcal{L}_{\text{policy}}(\xi) \!=\!
& -\mathbb{E}_{t}\Big[\min\Big( \rho_{t}(\xi)\hat{A}_{t}, \nonumber \\
& \quad\quad\;\;\, \mathrm{clip}\left(\rho_{t}(\xi),\, 1-\epsilon,\, 1+\epsilon\right)\hat{A}_{t} \Big)\Big],
\end{aligned}
\end{equation}
where the probability ratio $\rho_t(\xi)$ is:
\begin{equation}
\rho_t(\xi) = \frac{G(\xi)(a_t\mid s_t,m)}{G(\xi_{\text{old}})(a_t\mid s_t,m)},
\end{equation}
and $\epsilon$ is a clipping parameter.
The combined PPO loss to minimize is:
\begin{equation}
\mathcal{L}_{\text{PPO}}(\xi,\phi) = \mathcal{L}_{\text{policy}}(\xi) + c_{\text{VF}}\cdot \mathcal{L}_{\text{value}}(\phi),
\end{equation}
where $c_{\text{VF}}$ is the coefficient for the value objective. 

\section{Experiments}
\subsection{Dataset}
We conducted our experiments on the ESConv dataset, which contains 1,300 conversations comprising 38,365 utterances between help-seekers (user) and supporters. The help-seeker provides a feedback score for every two dialogue turns exchanged with the supporter. Accordingly, we concatenate each two-turn as a single input and pair it with the corresponding feedback score as the output. Each input also includes the preceding contexts among the same conversation. We split the dataset into training, validation, and test sets using an 8:1:1 ratio at the conversation level, ensuring no conversation overlap among these subsets.

\subsection{Settings}
All models were fine-tuned using parameter-efficient Low-Rank Adaptation (LoRA) \cite{hu2022lora} with 8-bit quantization. LoRA adapters were configured with a rank of $r=16$, scaling factor $\alpha=16$, and a dropout rate of $0.1$. The base model weights were frozen, and only LoRA parameters were updated. We fixed the random seed to 42 and set the maximum input length to 1024. At inference time, we applied top-$p$ sampling with $p=0.85$, a temperature of $0.7$, and a maximum generation length of 256 tokens.

During the SFT stage, we used a learning rate of $1\times10^{-4}$, a batch size of 8, and trained for up to 15 epochs with an early stopping patience of 3.
For the M\textsuperscript{2}PC stage, training was conducted for 10 EM iterations with a batch size of 2, gradient accumulation steps of 4, and a learning rate of $1\times10^{-5}$. We set diverse user clusters as 20 each for the majority and minority groups for clustering initialization.
Both SFT and M\textsuperscript{2}PC stages were trained on a single NVIDIA RTX A6000 GPU (49GB) using the AdamW optimizer \cite{loshchilovdecoupled}.

For the PAda-PPO stage, training was performed on 4 NVIDIA RTX A6000 GPUs (49GB each) using DeepSpeed ZeRO Stage 2 \cite{rasley2020deepspeed} and Hugging Face Accelerate \cite{accelerate}. We used a batch size of 2, gradient accumulation of 2, a learning rate of $3\times10^{-7}$, and trained for 5 epochs. Following \citet{ziegler2019fine, luong2024reft}, we set PPO hyperparameters as follows: $\lambda=1$, $\gamma=0.95$, $c_{\text{VF}}=0.1$, $\epsilon=0.2$. And the KL penalty coefficient is set to 0.2.

\subsection{Comparative Models}
We organized our experiments into three stages: \textit{zero-shot inference}, \textit{UCoT inference}, and \textit{supervised fine-tuning (SFT)} using LoRA.

\noindent\textbf{Zero-shot}: we evaluated three backbone models with base prompt (Appendix~\ref{sec:prompts_example}, Figure~\ref{fig:base_prompt}): Llama‑3.2‑1B-Instruct\footnote{\href{https://huggingface.co/meta-llama/Llama-3.2-1B-Instruct}{https://huggingface.co/meta-llama/Llama-3.2-1B-Instruct}}, Llama-3-8B-Instruct\footnote{\href{https://huggingface.co/meta-llama/Meta-Llama-3-8B-Instruct}{https://huggingface.co/meta-llama/Meta-Llama-3-8B-Instruct}}, and GPT4.1‑mini, without training.

\noindent \textbf{UCoT}: We augmented the input with UCoT prompts, without fine-tuning.

\noindent\textbf{SFT (LoRA)}: We fine-tuned Llama‑3.2‑1B-Instruct and Llama‑3‑8B-Instruct using LoRA adapters with both base and UCoT-augmented prompts.
While the model with UCoT or CoPeR module shares the same prompt, their training targets differ: the model with UCoT supervision uses only the final feedback score, whereas the model with CoPeR supervision additionally includes the synthesized step-by-step reasoning output ($r_{\text{coper}}$).

\noindent\textbf{Reinforcement Learning (RL)}: we further fine-tuned the SFT models using both standard PPO \cite{schulman2017proximal} and our proposed PAda‑PPO. Specifically, we applied these methods to the following SFT variants: Llama‑3‑8B-Instruct, Llama‑3‑8B-Instruct-UCoT, and Llama‑3‑8B-Instruct-CoPeR. 

\subsection{Evaluation Metrics}
We adopt four classification metrics: the $F_1$ score for each individual class (``low satisfaction'' and ``high satisfaction''), macro-averaged $F_1$, and weighted $F_1$.
The class-wise $F_1$ scores (denoted as $F_1^{\text{low}}$ and $F_1^{\text{high}}$ measure how well the model predicts each satisfaction label individually. The macro $F_1$ ($F_1^{\text{m}}$) is computed by averaging the $F_1$ scores of both classes, giving equal weight to each regardless of class frequency. The weighted $F_1$ ($F_1^{\text{w}}$) further accounts for label imbalance by weighting each class $F_1$ score by its support. 

\section{Results and Analysis}


\subsection{Evaluation of Synthesized Rationales by GPT-4.1-mini}
We evaluate the accuracy and quality of synthesized rationales by GPT-4.1-mini. Individual preference is inherently subjective, human evaluation may be unreliable: annotators cannot directly access the seeker’s real thoughts. 
We assess the quality and accuracy of the synthesized rationales with two objective metrics:

\noindent (1) \textbf{Supporter-strategy accuracy (7 classes)}: measures the alignment between the supporter strategy predicted in step 2 and the true label.

\noindent (2) \textbf{Logical accuracy}: measures the consistency between reasoning steps (Steps~1–3) and the final judgment (Step~4). Step~3 infers whether the supporter’s strategy aligns with the seeker's intent based on predictions from Steps~1–2, while Step~4 provides the gold satisfaction score (1–5). We map “matched intent” to scores 4–5, ``partially matched'' to 2–3, and ``did not match'' to 1. A rationale is considered logically correct when this mapping holds.
On the training set, the synthesized rationales by GPT-4.1-mini achieve 63.44\% supporter-strategy accuracy and 74.16\% logical accuracy; on the validation set, 61.04\% and 77.23\%, respectively. Appendix~\ref{sec:evaluation} presents additional case studies.

\begin{table}[h]
  \centering
  \scalebox{0.91}{
  \begin{tabular}{lcccc}
    \toprule
    Models & $F_1^{\text{low}}$ & $F_1^{\text{high}}$ & $F_1^{\text{w}}$ & $F_1^{\text{m}}$\\
    \midrule
    \multicolumn{5}{l}{\textit{Zero‑shot}}\\
    Llama‑3.2‑1B            & 0.27 & 0.49 & 0.45 & 0.38 \\
    Llama‑3‑8B              & 0.34 & 0.71 & 0.65 & 0.52 \\
    GPT4.1‑mini                      & \textbf{0.40} & 0.67 & 0.62 & 0.53 \\
    \midrule
    \multicolumn{5}{l}{\textit{\textbf{UCoT}}}\\
    Llama‑3.2‑1B-UCoT       & 0.28 & 0.40 & 0.38 & 0.34 \\
    Llama‑3‑8B-UCoT         & 0.26 & 0.80 & 0.72 & 0.53 \\
    GPT4.1‑mini-UCoT                 & 0.38 & 0.57 & 0.54 & 0.47 \\
    \midrule
    \multicolumn{5}{l}{\textit{SFT (with LoRA)}}\\
    Llama‑3.2‑1B            & 0.28 & 0.76 & 0.67 & 0.52 \\
    Llama‑3.2‑1B-\textbf{UCoT}       & 0.18 & \textbf{0.88} & 0.75 & 0.53 \\
    Llama‑3.2‑1B-\textbf{CoPeR}       & 0.27  & 0.85  & 0.74  & 0.56 \\
    Llama‑3‑8B              & 0.24 & 0.82 & 0.71 & 0.53 \\
    Llama‑3‑8B-\textbf{UCoT}         & 0.27 & 0.86 & 0.75 & 0.56 \\
    Llama‑3‑8B-\textbf{CoPeR} & 0.30 & 0.86 & \textbf{0.76} & \textbf{0.58} \\
    \bottomrule
      \end{tabular}}
    \caption{\textbf{Results of the proposed UCoT and CoPeR:} Comparison of baseline models and our proposed UCoT and CoPeR module. We report $F_1$ scores on low ($F_1^{\text{low}}$) and high ($F_1^{\text{high}}$) satisfaction classes, along with weighted ($F_1^{\text{w}}$) and macro ($F_1^{\text{m}}$) averages. \textbf{Bold} in $F_1$ indicates the best performance. All Llama models refer to the instruction-tuned versions.}
  \label{tab:sft}
\end{table}

\begin{table*}[t]
\centering
\scalebox{0.92}{
\begin{tabular}{lcccccccccccc}
\toprule
\multirow{2}{*}{Models} & \multicolumn{4}{c}{Minority} & \multicolumn{4}{c}{Majority} & \multicolumn{4}{c}{Minority+Majority} \\
\cmidrule(lr){2-5} \cmidrule(lr){6-9} \cmidrule(lr){10-13}
                        & $F_1^{\text{low}}$ & $F_1^{\text{high}}$ &$F_1^{\text{w}}$ & $F_1^{\text{m}}$ & $F_1^{\text{low}}$ & $F_1^{\text{high}}$ & $F_1^{\text{w}}$ & $F_1^{\text{m}}$ &$F_1^{\text{low}}$ & $F_1^{\text{high}}$ & $F_1^{\text{w}}$ & $F_1^{\text{m}}$ \\
\midrule
Llama-3-8B         & 0.17 & 0.47 & 0.28 & 0.32 & 0.16 & 0.90 & 0.84 & 0.53 & 0.16 & 0.84 & 0.71 & 0.50 \\
\quad +\textbf{M\textsuperscript{2}PC}                  & 0.68 & \textbf{0.55} & 0.61 & 0.61 & 0.19 & 0.94 & \textbf{0.89} & \textbf{0.57} & 0.56 & 0.89 & 0.82 & 0.72 \\
\midrule
Llama-3-8B-UCoT    & 0.36 & 0.49 & 0.41 & 0.42 & 0.21 & 0.88 & 0.83 & 0.55 & 0.26 & 0.83 & 0.72 & 0.55 \\
\quad +\textbf{M\textsuperscript{2}PC}                  & 0.73 & 0.18 & 0.56 & 0.46 & 0.00 & \textbf{0.96} & 0.88 & 0.48 & 0.58 & 0.92 & 0.85 & 0.75 \\
\midrule
Llama-3-8B-CoPeR  & 0.29 & 0.53 & 0.37 & 0.41 & \textbf{0.21} & 0.90 & 0.85 & 0.56 & 0.24 & 0.85 & 0.74 & 0.55 \\
\quad +\textbf{M\textsuperscript{2}PC}                    & \textbf{0.80} & 0.46 & \textbf{0.70} & \textbf{0.63} & 0.08 & 0.95 & 0.87 & 0.51 & \textbf{0.60} & \textbf{0.92} & \textbf{0.86} & \textbf{0.76} \\
\bottomrule
\end{tabular}
}
\caption{\textbf{Results of the proposed M\textsuperscript{2}PC:} $F_1$ scores on the validation set across three model backbones, evaluated on minority, majority, and combined user groups. For each model, the reported M\textsuperscript{2}PC results correspond to the best-performing EM iteration.  
Llama‑3‑8B refers to the Llama‑3‑8B‑Instruct version.}
\label{tab:em}
\end{table*}

\subsection{Results of the Proposed UCoT and CoPeR}
Table~\ref{tab:sft} reports classification performance across three settings: \textit{Zero-shot}, \textit{UCoT}, and \textit{SFT (with LoRA)}. Overall, supervised fine-tuning (SFT) consistently outperforms both the zero-shot and inference-only UCoT settings, emphasizing the importance of task-specific adaptation. Prompting with UCoT alone results in degraded performance compared to the zero-shot baseline, suggesting that, without a supervision, UCoT may overly constrain the model's reasoning process, failing to predict correctly.
Within the SFT setting, models fine-tuned with UCoT prompts consistently outperform their counterparts across both backbone architectures (Llama-3.2-1B-Instruct and Llama-3-8B-Instruct). Furthermore, our proposed CoPeR approach, which extends UCoT by additionally supervising the model to generate user-specific reasoning as output, achieves the highest performance in both weighted and macro $F_1$ scores. These results highlight the effectiveness of combining user-specific prompting with explicit supervision of personalized reasoning in enhancing the model's ability to predict user satisfaction from dialogue.


\subsection{Results of the Proposed M\textsuperscript{2}PC}
\subsubsection{Analysis of Majority and Minority}
We first evaluate the proposed method on the majority and minority groups to assess its group-level performance. Table~\ref{tab:em} reports $F_1$ scores on the validation set across three model backbones: Llama-3-8B-Instruct, Llama-3-8B-Instruct-UCoT, and Llama-3-8B-Instruct-CoPeR, before and after applying our proposed M\textsuperscript{2}PC method. Across all configurations, M\textsuperscript{2}PC consistently improves overall performance, as measured by all four metrics on the combined minority and majority groups. For example, when applied to Llama-3-8B-Instruct-CoPeR, M\textsuperscript{2}PC improves $F_1^{\text{low}}$ from 0.24 to 0.60, $F_1^{\text{high}}$ from 0.85 to 0.92, $F_1^{\text{w}}$ from 0.74 to 0.86, and $F_1^{\text{m}}$ from 0.55 to 0.76.
For the minority group, M\textsuperscript{2}PC leads to significant improvements in low satisfaction prediction across all backbones.  In the CoPeR setting, for example, $F_1^{\text{low}}$ increases from 0.29 to 0.80, $F_1^{\text{w}}$ from 0.37 to 0.70, and $F_1^{\text{m}}$ from 0.41 to 0.63. For the majority group, M\textsuperscript{2}PC consistently enhances high satisfaction prediction as well, for example, increasing from 0.90 to 0.95 in the CoPeR configuration. Although there is a modest decline in high satisfaction scores for the minority group and low satisfaction scores for the majority group, the overall performance improves, as reflected in higher weighted $F_1$ scores. The detailed results and trends across EM iterations can be seen in Appendix~\ref{sec:em_iterations}.

These results highlight the effectiveness of modeling group-specific preferences through clustering. M\textsuperscript{2}PC successfully adapts to diverse user satisfaction patterns by routing each user to the model most aligned with their latent preference semantics, enabling more accurate prediction across minority-majority groups. We further investigate the effect of different initializations of user clusters for majority and minority groups, with details provided in Appendix~\ref{sec:ablation}.

\subsubsection{Analysis of Subgroups Within Majority and Minority}
Next, we analyze finer-grained subgroups within the majority and minority groups. We applied the proposed M\textsuperscript{2}PC method to cluster the validation set into majority and minority groups. Within each group, we extracted the last hidden states from the corresponding (majority or minority) model outputs and applied $k$-means++ \cite{arthur2007k} to cluster users into 2-20 subgroups. We then used the silhouette score \cite{rousseeuw1987silhouettes} to determine the optimal number of clusters ($k$) and computed the weighted $F_1$ for each cluster ($\in k$). 

Table~\ref{tab:k_subgroups} shows that our method captures most user subgroups, including smaller yet distinct clusters (in bold), suggesting that M\textsuperscript{2}PC can generally adapt to diverse intra-group preferences. 
Since weighted $F_1$ is affected by the model's overall performance, we adopt a relative criterion: if smaller subgroups outperform the two largest ones in weighted $F_1$ within either the majority or minority population, it indicates that the model capture diverse subgroup characteristics rather than overfitting to the more frequent patterns.

\begin{table*}[t]
\centering
\resizebox{\textwidth}{!}{
\begin{tabular}{lllllllllllll}
\toprule
Models & 1 & 2 & 3 & 4 & 5 & 6 & 7 & 8 & 9 & 10 \\
\midrule
\multicolumn{11}{l}{Llama-3-8B+M\textsuperscript{2}PC} \\
Maj. & \textit{0.85(111)} & \textit{0.68(105)} & \textbf{0.89(67)} & \textbf{0.84(60)} & \textbf{0.91(55)} & \textbf{0.91(38)} & \textbf{0.96(38)} & \textbf{0.92(31)} & \textbf{0.95(27)} & \textbf{0.79(21)} \\
Min. & \textit{0.60(21)} & \textit{0.71(21)} & \textbf{0.68(10)} & \textbf{0.77(9)} & 0.63(8) & 0.19(7) & \textbf{0.84(6)} & \textbf{0.91(6)} & 0.57(5) & \textbf{1.00(5)} \\
\midrule
\multicolumn{11}{l}{Llama-3-8B-UCoT+M\textsuperscript{2}PC} \\
Maj. & \textit{0.68(109)} & \textit{0.82(104)} & \textbf{0.94(50)} & \textbf{0.87(44)} & \textbf{0.96(42)} & \textbf{0.93(41)} & \textbf{0.86(32)} & \textbf{1.00(31)} & \textbf{0.89(28)} & \textbf{1.00(24)} \\
Min. & \textit{0.62(29)} & \textit{0.61(21)} & 0.60(11) & \textbf{0.87(8)} & \textbf{0.91(6)} & \textbf{0.91(6)} & \textbf{0.78(5)} & \textbf{0.80(5)} & \textbf{0.73(4)} & \textbf{1.00(4)} \\
\midrule
\multicolumn{11}{l}{Llama-3-8B-CoPeR+M\textsuperscript{2}PC} \\
Maj. & \textit{0.71(134)} & \textit{0.79(105)} & \textbf{0.93(56)} & \textbf{0.85(48)} & \textbf{0.94(44)} & \textbf{0.94(39)} & \textbf{0.90(30)} & \textbf{0.90(30)} & \textbf{0.82(28)} & \textbf{0.87(27)} \\
Min. & \textit{0.70(22)} & \textit{0.61(21)} & \textbf{0.67(7)} & \textbf{0.91(6)} & \textbf{0.67(6)} & \textbf{1.00(5)} & \textbf{0.80(5)} & 0.34(5) & 0.64(4) & 0.33(4) \\
\midrule
\end{tabular}
}
\end{table*}

\begin{table*}[h]
\centering
\resizebox{\textwidth}{!}{
\begin{tabular}{lllllllllllll}
\midrule
Models & 11 & 12 & 13 & 14 & 15 & 16 & 17 & 18 & 19 & 20 \\
\midrule
\multicolumn{11}{l}{Llama-3-8B+M\textsuperscript{2}PC} \\
Maj. & \textbf{1.00(18)} & \textbf{0.91(17)} & \textbf{1.00(10)} & \textbf{0.84(9)} & \textbf{1.00(9)} & \textbf{1.00(9)} & \textbf{1.00(6)} & - & - & - \\
Min. & \textbf{0.82(5)} & 0.63(5) & 0.00(4) & 0.64(4) & 0.50(4) & \textbf{0.73(4)} & \textbf{0.73(4)} & \textbf{0.73(4)} & \textbf{0.67(3)} & - \\
\midrule
\multicolumn{11}{l}{Llama-3-8B-UCoT+M\textsuperscript{2}PC} \\
Maj. & \textbf{0.88(24)} & \textbf{0.87(22)} & \textbf{0.90(15)} & \textbf{1.00(14)} & \textbf{1.00(13)} & \textbf{1.00(12)} & \textbf{1.00(9)} & \textbf{1.00(8)} & \textbf{1.00(5)} & \textbf{1.00(3)} \\
Min. & 0.33(4) & \textbf{1.00(4)} & 0.10(4) & 0.33(3) & 0.53(3) & 0.53(3) & \textbf{0.67(2)} & \textbf{0.67(2)} & 0.00(2) & 0.00(2)   \\
\midrule
\multicolumn{11}{l}{Llama-3-8B-CoPeR+M\textsuperscript{2}PC} \\
Maj. & \textbf{0.91(22)} & \textbf{0.94(20)} & \textbf{0.90(19)} & \textbf{0.96(14)} & \textbf{1.00(9)} & \textbf{1.00(8)} & \textbf{1.00(8)} & \textbf{1.00(7)} & \textbf{1.00(4)} & - \\
Min. & \textbf{1.00(4)} & 0.20(4) & \textbf{0.67(3)} & 0.53(3) & 0.53(3) & \textbf{0.67(3)} & \textbf{1.00(3)} & \textbf{0.67(2)} & \textbf{1.00(2)} & \textbf{1.00(2)} \\
\bottomrule
\end{tabular}
}
\caption{Weighted $F_1$ scores across $k \in {1,\ldots,20}$ for the majority (Maj.) and minority (Min.) user groups, shown in two parts for readability (top: $k=1$–$10$; bottom: $k=11$–$20$). Numbers in parentheses indicate the number of users in each subgroup, where subgroups ($k=1$–$20$) are ordered by their user counts. Bold values denote subgroups whose weighted $F_1$ exceeds the average of the two largest subgroups (shown in \textit{italic}).}
\label{tab:k_subgroups}
\end{table*}

\begin{table}[h]
  \centering
  \scalebox{0.95}{
  \begin{tabular}{lcccc}
    \toprule
    Models & $F_1^{\text{low}}$ & $F_1^{\text{high}}$ & $F_1^{\text{w}}$ & $F_1^{\text{m}}$ \\
    \midrule
    Llama‑3‑8B
    & 0.24 & 0.82 & 0.71 & 0.53 \\
    \quad + PPO                          & 0.25 & 0.85 & 0.74 & 0.55 \\
    \quad + \textbf{PAda‑PPO}                      & 0.29 & 0.86 & 0.75 & 0.57 \\
    \midrule
    Llama‑3‑8B-UCoT             & 0.27 & 0.86 & 0.75 & 0.56 \\
    \quad + PPO                          & 0.22 & 0.88 & 0.76 & 0.55 \\
    \quad + \textbf{PAda‑PPO}                      & \textbf{0.36} & 0.86 & 0.77 & \textbf{0.61} \\
    \midrule
    Llama‑3‑8B-CoPeR           & 0.30 & 0.86 & 0.76 & 0.58 \\
    \quad + PPO                          & 0.34 & \textbf{0.88} & \textbf{0.78} & 0.61\\
    \quad + \textbf{PAda‑PPO}                      & 0.33   & 0.85   & 0.76   & 0.59   \\ 
    \bottomrule
      \end{tabular}}
      \caption{\textbf{Results of the proposed PAda-PPO:} Comparison among SFT, PPO-based RL, and our proposed PAda-PPO method. Llama‑3‑8B refers to the Llama‑3‑8B‑Instruct version.}
    \label{tab:adp_ppo}
\end{table}

\subsection{Results of the Proposed PAda-PPO}
Table~\ref{tab:adp_ppo} presents the performance of our proposed PAda-PPO algorithm applied to three model backbones: Llama‑3‑8B-Instruct (base), Llama‑3‑8B-Instruct-UCoT, and Llama‑3‑8B-Instruct-CoPeR. We compare three strategies: SFT, SFT followed by PPO, and SFT followed by PAda-PPO. Across both the base and UCoT-augmented backbones, PAda-PPO consistently outperforms SFT and PPO. For instance, with Llama‑3‑8B-Instruct-UCoT, PAda-PPO outperforms PPO on $F_1^{\text{low}}$ from 0.22 to 0.36 and macro $F_1$ from 0.55 to 0.61, demonstrating its effectiveness in improving minority-class satisfaction prediction while maintaining strong performance on the majority class.

However, on the Llama‑3‑8B-Instruct-CoPeR backbone, PAda-PPO performs slightly worse than PPO. We attribute this to the noise introduced by the GPT‑4.1‑mini synthesized CoPeR rationales. When the synthesized user-specific reasoning conflicts with the preference routing signals, it will lead to a higher KL divergence penalty (Equation~\ref{eq:kl}), destabilizing training. Improving the quality of CoPeR synthesis will be done in future work.

\section{Conclusions and Future Work}
We have introduced a unified framework for satisfaction estimation that simultaneously models individual- and group-level preferences among majority and minority user populations. We proposed UCoT and CoPeR to generate interpretable reasoning chains for capturing individual preferences, and developed M\textsuperscript{2}PC, an unsupervised clustering module for identifying group-level preferences. These were integrated into PAda-PPO to align dialogue systems with diverse user preferences. Experiments on the ESConv dataset demonstrate improved satisfaction estimation across different user populations.
In future work, we will validate our proposed method on other LLMs, 
and extend preference-adaptive reinforcement learning to additional RL algorithms. 

\section*{Limitations}
A key limitation of our approach lies in the CoPeR supervision stage, where we rely on GPT‑4.1‑mini to generate reasoning chains as pseudo-ground truth. Given the inherently subjective nature of user satisfaction and context interpretation, it is challenging to synthesize rationales that accurately reflect the user's underlying intent. While this setup enables scalable training without manual annotation, the quality of the generated rationales may affect downstream fine-tuning and reinforcement learning. Inaccurate or overly generic reasoning can introduce noise into the learning process, hindering the model's ability to capture nuanced satisfaction patterns. 
Future work could collect partial real human data, including users' own intents and explanations for their feedback scores, to enhance the reliability of reasoning supervision.

Another limitation is that M\textsuperscript{2}PC does not model some subgroups of minority users well, as shown in Table~\ref{tab:k_subgroups}. Future work is needed to better account for the nuanced preference patterns among these underrepresented subgroups.

\section*{Acknowledgements}
This work was supported by JST Moonshot R\&D Goal 1 Avatar Symbiotic Society Project (JPMJMS2011). 

\bibliography{custom}

@inproceedings{liu2021towards,
  title={Towards Emotional Support Dialog Systems},
  author={Liu, Siyang and Zheng, Chujie and Demasi, Orianna and Sabour, Sahand and Li, Yu and Yu, Zhou and Jiang, Yong and Huang, Minlie},
  booktitle={Proceedings of ACL-IJCNLP},
  pages={3469--3483},
  year={2021}
}

@article{wang2023aligning,
  title={Aligning language models with human preferences via a bayesian approach},
  author={Wang, Jiashuo and Wang, Haozhao and Sun, Shichao and Li, Wenjie},
  journal={Advances in Neural Information Processing Systems},
  volume={36},
  pages={49113--49132},
  year={2023}
}

@article{chakraborty2024maxmin,
  title={MaxMin-RLHF: Alignment with Diverse Human Preferences},
  author={Chakraborty, Souradip and Qiu, Jiahao and Yuan, Hui and Koppel, Alec and Manocha, Dinesh and Huang, Furong and Bedi, Amrit Singh and Wang, Mengdi},
  journal={Proceedings of Machine Learning Research},
  volume={235},
  pages={6116--6135},
  year={2024},
  publisher={ML Research Press}
}

@article{jang2023personalized,
  title={Personalized Soups: Personalized Large Language Model Alignment via Post-hoc Parameter Merging},
  author={Jang, Joel and Kim, Seungone and Lin, Bill Yuchen and Wang, Yizhong and Hessel, Jack and Zettlemoyer, Luke and Hajishirzi, Hannaneh and Choi, Yejin and Ammanabrolu, Prithviraj},
  journal={arXiv preprint arXiv:2310.11564},
  year={2023}
}

@article{xiao2024algorithmic,
  title={On the algorithmic bias of aligning large language models with RLHF: Preference collapse and matching regularization},
  author={Xiao, Jiancong and Li, Ziniu and Xie, Xingyu and Getzen, Emily and Fang, Cong and Long, Qi and Su, Weijie J},
  journal={arXiv preprint arXiv:2405.16455},
  year={2024}
}

@inproceedings{slocum2025diverse,
  title={Diverse preference learning for capabilities and alignment},
  author={Slocum, Stewart and Parker-Sartori, Asher and Hadfield-Menell, Dylan},
  booktitle={The Thirteenth International Conference on Learning Representations},
  year={2025}
}

@inproceedings{yang2024llm,
  title={Llm voting: Human choices and ai collective decision-making},
  author={Yang, Joshua C et al.},
  booktitle={Proceedings of the AAAI/ACM Conference on AI, Ethics, and Society},
  volume={7},
  pages={1696--1708},
  year={2024}
}

@article{li2024personalized,
  title={Personalized language modeling from personalized human feedback},
  author={Li, Xinyu and Zhou, Ruiyang and Lipton, Zachary C and Leqi, Liu},
  journal={arXiv preprint arXiv:2402.05133},
  year={2024}
}

@article{xu2024multi,
  title={Multi-dimensional evaluation of empathetic dialog responses},
  author={Xu, Zhichao and Jiang, Jiepu},
  journal={arXiv preprint arXiv:2402.11409},
  year={2024}
}

@inproceedings{lin-etal-2024-interpretable,
    title = "Interpretable User Satisfaction Estimation for Conversational Systems with Large Language Models",
    author={Lin, Ying-Chun and Neville, Jennifer and Stokes, Jack and Yang, Longqi and Safavi, Tara and Wan, Mengting and Counts, Scott and Suri, Siddharth and Andersen, Reid and Xu, Xiaofeng and others},
    booktitle = "Proceedings of the 62nd Annual Meeting of the Association for Computational Linguistics (Volume 1: Long Papers)",
    month = aug,
    year = "2024",
    url = "https://aclanthology.org/2024.acl-long.598/",
    doi = "10.18653/v1/2024.acl-long.598",
    pages = "11100--11115"
}

@article{yang2024rewards,
  title={Rewards-in-context: Multi-objective alignment of foundation models with dynamic preference adjustment},
  author={Yang, Rui and Pan, Xiaoman and Luo, Feng and Qiu, Shuang and Zhong, Han and Yu, Dong and Chen, Jianshu},
  journal={arXiv preprint arXiv:2402.10207},
  year={2024}
}

@article{wang2024arithmetic,
  title={Arithmetic control of llms for diverse user preferences: Directional preference alignment with multi-objective rewards},
  author={Wang, Haoxiang and Lin, Yong and Xiong, Wei and Yang, Rui and Diao, Shizhe and Qiu, Shuang and Zhao, Han and Zhang, Tong},
  journal={arXiv preprint arXiv:2402.18571},
  year={2024}
}

@article{luong2024reft,
  title={Reft: Reasoning with reinforced fine-tuning},
  author={Luong, Trung Quoc and Zhang, Xinbo and Jie, Zhanming and Sun, Peng and Jin, Xiaoran and Li, Hang},
  journal={arXiv preprint arXiv:2401.08967},
  year={2024}
}

@article{la2025fairness,
  title={Fairness Aware Reinforcement Learning via Proximal Policy Optimization},
  author={La Malfa, Gabriele and Zhang, Jie M and Luck, Michael and Black, Elizabeth},
  journal={arXiv preprint arXiv:2502.03953},
  year={2025}
}

@article{wei2022chain,
  title={Chain-of-thought prompting elicits reasoning in large language models},
  author={Wei, Jason and Wang, Xuezhi and Schuurmans, Dale and Bosma, Maarten and Xia, Fei and Chi, Ed and Le, Quoc V and Zhou, Denny and others},
  journal={Advances in neural information processing systems},
  volume={35},
  pages={24824--24837},
  year={2022}
}

@article{fu2023dual,
  title={Dual variational generative model and auxiliary retrieval for empathetic response generation by conversational robot},
  author={Fu, Yahui and Inoue, Koji and Lala, Divesh and Yamamoto, Kenta and Chu, Chenhui and Kawahara, Tatsuya},
  journal={Advanced Robotics},
  volume={37},
  number={21},
  pages={1406--1418},
  year={2023},
  publisher={Taylor \& Francis}
}

@inproceedings{fu2023reasoning,
  title={Reasoning before Responding: Integrating Commonsense-based Causality Explanation for Empathetic Response Generation},
  author={Fu, Yahui and Inoue, Koji and Chu, Chenhui and Kawahara, Tatsuya},
  booktitle={Proceedings of the 24th Annual Meeting of the Special Interest Group on Discourse and Dialogue},
  pages={645--656},
  year={2023}
}

@inproceedings{zhang2024escot,
  title={ESCoT: Towards Interpretable Emotional Support Dialogue Systems},
  author={Zhang, Tenggan and Zhang, Xinjie and Zhao, Jinming and Zhou, Li and Jin, Qin},
  booktitle={Proceedings of the 62nd Annual Meeting of the Association for Computational Linguistics (Volume 1: Long Papers)},
  pages={13395--13412},
  year={2024}
}

@inproceedings{fu2024styemp,
  title={StyEmp: Stylizing Empathetic Response Generation via Multi-Grained Prefix Encoder and Personality Reinforcement},
  author={Fu, Yahui and Chu, Chenhui and Kawahara, Tatsuya},
  booktitle={Proceedings of the 25th Annual Meeting of the Special Interest Group on Discourse and Dialogue},
  pages={172--185},
  year={2024}
}

@inproceedings{xie2025leveraging,
  title={Leveraging chain of thought towards empathetic spoken dialogue without corresponding question-answering data},
  author={Xie, Jingran and Lei, Shun and Yu, Yue and Xiang, Yang and Wang, Hui and Wu, Xixin and Wu, Zhiyong},
  booktitle={ICASSP},
  pages={1--5},
  year={2025},
  organization={IEEE}
}

@article{aroyo2023dices,
  title={Dices dataset: Diversity in conversational ai evaluation for safety},
  author={Aroyo, Lora and Taylor, Alex and Diaz, Mark and Homan, Christopher and Parrish, Alicia and Serapio-Garc{\'\i}a, Gregory and Prabhakaran, Vinodkumar and Wang, Ding},
  journal={Advances in Neural Information Processing Systems},
  volume={36},
  pages={53330--53342},
  year={2023}
}

@article{costa2014associations,
  title={Associations between medical student empathy and personality: a multi-institutional study},
  author={Costa, Patricio and Alves, Raquel and Neto, Isabel and Marvao, Pedro and Portela, Miguel and Costa, Manuel Joao},
  journal={PloS one},
  volume={9},
  number={3},
  pages={e89254},
  year={2014},
  publisher={Public Library of Science San Francisco, USA}
}

@article{richendoller1994exploring,
  title={Exploring the links between personality and empathic response style},
  author={Richendoller, Nadine R and Weaver III, James B},
  journal={Personality and individual Differences},
  volume={17},
  number={3},
  pages={303--311},
  year={1994},
  publisher={Elsevier}
}

@article{schulman2017proximal,
  title={Proximal policy optimization algorithms},
  author={Schulman, John and Wolski, Filip and Dhariwal, Prafulla and Radford, Alec and Klimov, Oleg},
  journal={arXiv preprint arXiv:1707.06347},
  year={2017}
}

@article{schulman2015high,
  title={High-dimensional continuous control using generalized advantage estimation},
  author={Schulman, John and Moritz, Philipp and Levine, Sergey and Jordan, Michael and Abbeel, Pieter},
  journal={arXiv preprint arXiv:1506.02438},
  year={2015}
}

@article{trott2019keeping,
  title={Keeping your distance: Solving sparse reward tasks using self-balancing shaped rewards},
  author={Trott, Alexander and Zheng, Stephan and Xiong, Caiming and Socher, Richard},
  journal={Advances in Neural Information Processing Systems},
  volume={32},
  year={2019}
}

@article{zheng2023secrets,
  title={Secrets of rlhf in large language models part i: Ppo},
  author={Zheng, Rui and Dou, Shihan and Gao, Songyang and Hua, Yuan and Shen, Wei and Wang, Binghai and Liu, Yan and Jin, Senjie and Liu, Qin and Zhou, Yuhao and others},
  journal={arXiv preprint arXiv:2307.04964},
  year={2023}
}

@article{kullback1951information,
  title={On information and sufficiency},
  author={Kullback, Solomon and Leibler, Richard A},
  journal={The annals of mathematical statistics},
  volume={22},
  number={1},
  pages={79--86},
  year={1951},
  publisher={JSTOR}
}

@article{ziegler2019fine,
  title={Fine-tuning language models from human preferences},
  author={Ziegler, Daniel M and Stiennon, Nisan and Wu, Jeffrey and Brown, Tom B and Radford, Alec and Amodei, Dario and Christiano, Paul and Irving, Geoffrey},
  journal={arXiv preprint arXiv:1909.08593},
  year={2019}
}

@inproceedings{rasley2020deepspeed,
  title={Deepspeed: System optimizations enable training deep learning models with over 100 billion parameters},
  author={Rasley, Jeff and Rajbhandari, Samyam and Ruwase, Olatunji and He, Yuxiong},
  booktitle={Proceedings of the 26th ACM SIGKDD international conference on knowledge discovery \& data mining},
  pages={3505--3506},
  year={2020}
}

@Misc{accelerate,
  title =        {Accelerate: Training and inference at scale made simple, efficient and adaptable.},
  author =       {Sylvain Gugger and Lysandre Debut and Thomas Wolf and Philipp Schmid and Zachary Mueller and Sourab Mangrulkar and Marc Sun and Benjamin Bossan},
  howpublished = {\url{https://github.com/huggingface/accelerate}},
  year =         {2022}
}

@inproceedings{loshchilovdecoupled,
  title={Decoupled Weight Decay Regularization},
  author={Loshchilov, Ilya and Hutter, Frank},
  booktitle={International Conference on Learning Representations},
year={2017}
}

@article{hu2022lora,
  title={Lora: Low-rank adaptation of large language models.},
  author={Hu, Edward J and Shen, Yelong and Wallis, Phillip and Allen-Zhu, Zeyuan and Li, Yuanzhi and Wang, Shean and Wang, Lu and Chen, Weizhu and others},
  journal={ICLR},
  volume={1},
  number={2},
  pages={3},
  year={2022}
}

@inproceedings{bert-score,
  title={BERTScore: Evaluating Text Generation with BERT},
  author={Tianyi Zhang and Varsha Kishore and Felix Wu and Kilian Q. Weinberger and Yoav Artzi},
  booktitle={International Conference on Learning Representations},
  year={2020},
  url={https://openreview.net/forum?id=SkeHuCVFDr}
}

@article{dempster1977maximum,
  title={Maximum likelihood from incomplete data via the EM algorithm},
  author={Dempster, Arthur P and Laird, Nan M and Rubin, Donald B},
  journal={Journal of the royal statistical society: series B (methodological)},
  volume={39},
  number={1},
  pages={1--22},
  year={1977},
  publisher={Wiley Online Library}
}

@article{touvron2023llama,
  title={Llama 2: Open foundation and fine-tuned chat models},
  author={Touvron, Hugo and Martin, Louis and Stone, Kevin and Albert, Peter and Almahairi, Amjad and Babaei, Yasmine and Bashlykov, Nikolay and Batra, Soumya and Bhargava, Prajjwal and Bhosale, Shruti and others},
  journal={arXiv preprint arXiv:2307.09288},
  year={2023}
}

@inproceedings{chae2023dialogue,
  title={Dialogue Chain-of-Thought Distillation for Commonsense-aware Conversational Agents},
  author={Chae, Hyungjoo and Song, Yongho and Ong, Kai Tzu Iunn and Kwon, Taeyoon and Kim, Minjin and Yu, Youngjae and Lee, Dongha and Kang, Dongyeop and Yeo, Jinyoung},
  booktitle={2023 Conference on Empirical Methods in Natural Language Processing, EMNLP 2023},
  pages={5606--5632},
  year={2023},
}

@article{rousseeuw1987silhouettes,
  title={Silhouettes: a graphical aid to the interpretation and validation of cluster analysis},
  author={Rousseeuw, Peter J},
  journal={Journal of computational and applied mathematics},
  volume={20},
  pages={53--65},
  year={1987},
  publisher={Elsevier}
}

@inproceedings{arthur2007k,
  title={k-means++ the advantages of careful seeding},
  author={Arthur, David and Vassilvitskii, Sergei},
  booktitle={Proceedings of the eighteenth annual ACM-SIAM symposium on Discrete algorithms},
  pages={1027--1035},
  year={2007}
}

@inproceedings{song-etal-2019-using,
  title     = {Using Customer Service Dialogues for Satisfaction Analysis with Context-Assisted Multiple Instance Learning},
  author    = {Song, Kaiqiang and Li, Xiaoyang and Zhang, Xiangchen and Yu, Dong and Carin, Lawrence},
  booktitle = {Proceedings of EMNLP-IJCNLP},
  year      = {2019},
  url       = {https://aclanthology.org/D19-1019/}
}

@inproceedings{deng-etal-2022-user,
  title     = {User Satisfaction Estimation with Sequential Dialogue Act Modeling in Goal-oriented Conversational Systems},
  author    = {Deng, Yang and Zhang, Wenxuan and Lam, Wai and Cheng, Hong and Meng, Helen},
  booktitle = {Proceedings of The Web Conference (WWW)},
  year      = {2022},
  url       = {https://arxiv.org/abs/2202.02912}
}

@inproceedings{ye2023modeling,
  title={Modeling User Satisfaction Dynamics in Dialogue via Hawkes Process},
  author={Ye, Fanghua and Hu, Zhiyuan and Yilmaz, Emine},
  booktitle={Proceedings of the 61st Annual Meeting of the Association for Computational Linguistics (Volume 1: Long Papers)},
  pages={8875--8889},
  year={2023}
}

@inproceedings{choi2019convSAT,
  title     = {Offline and Online Satisfaction Prediction in Open-Domain Conversational Systems},
  author    = {Choi, Jason Ingyu and Ahmadvand, Ali and Agichtein, Eugene},
  booktitle = {Proceedings of CIKM},
  year      = {2019},
  url       = {https://arxiv.org/abs/2006.01921}
}

@inproceedings{kim-etal-2025-llm,
  title     = {{LLM}-guided Plan and Retrieval: A Strategic Alignment for Interpretable User Satisfaction Estimation in Dialogue},
  author    = {Kim, Sangyeop and Park, Sohhyung and Jung, Jaewon and Kim, Jinseok and Cho, Sungzoon},
  booktitle = {Proceedings of NAACL},
  year      = {2025},
  url       = {https://aclanthology.org/2025.naacl-long.523.pdf}
}

@inproceedings{see2021understanding,
  title={Understanding and predicting user dissatisfaction in a neural generative chatbot},
  author={See, Abigail and Manning, Christopher D},
  booktitle={Proceedings of the 22nd Annual Meeting of the Special Interest Group on Discourse and Dialogue},
  pages={1--12},
  year={2021}
}

\appendix

\section{Prompts}
\label{sec:prompts_example}

\begin{figure*}
    \centering
\includegraphics[width=0.93\textwidth]{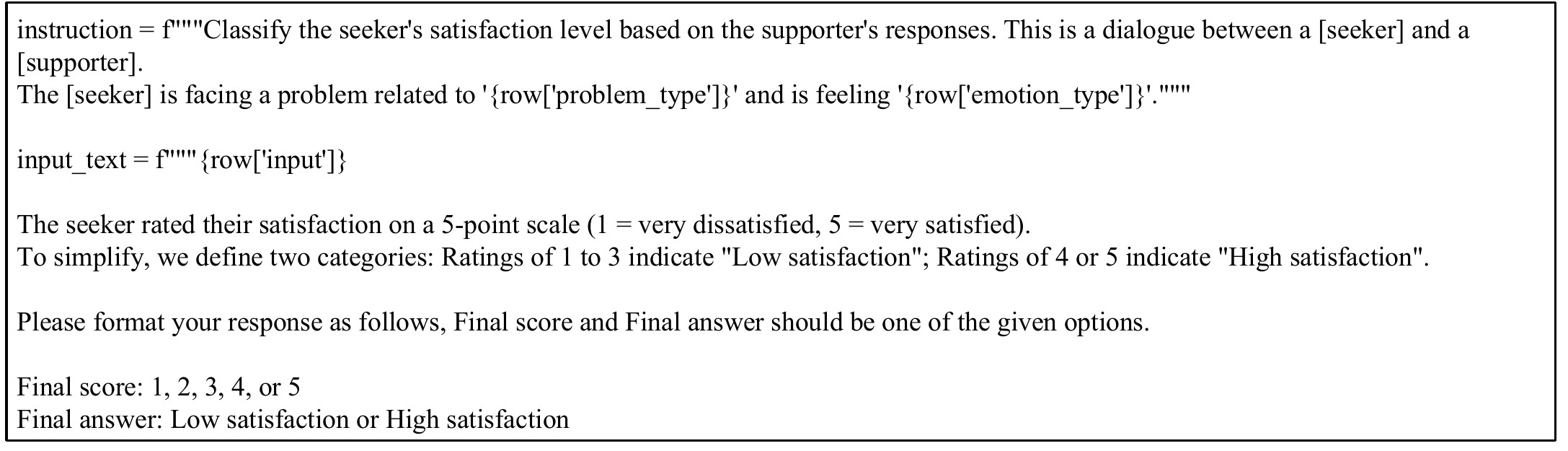}
    \caption{Base prompt.}
    \label{fig:base_prompt}
\end{figure*}

\begin{figure*}
    \centering
\includegraphics[width=0.93\textwidth]{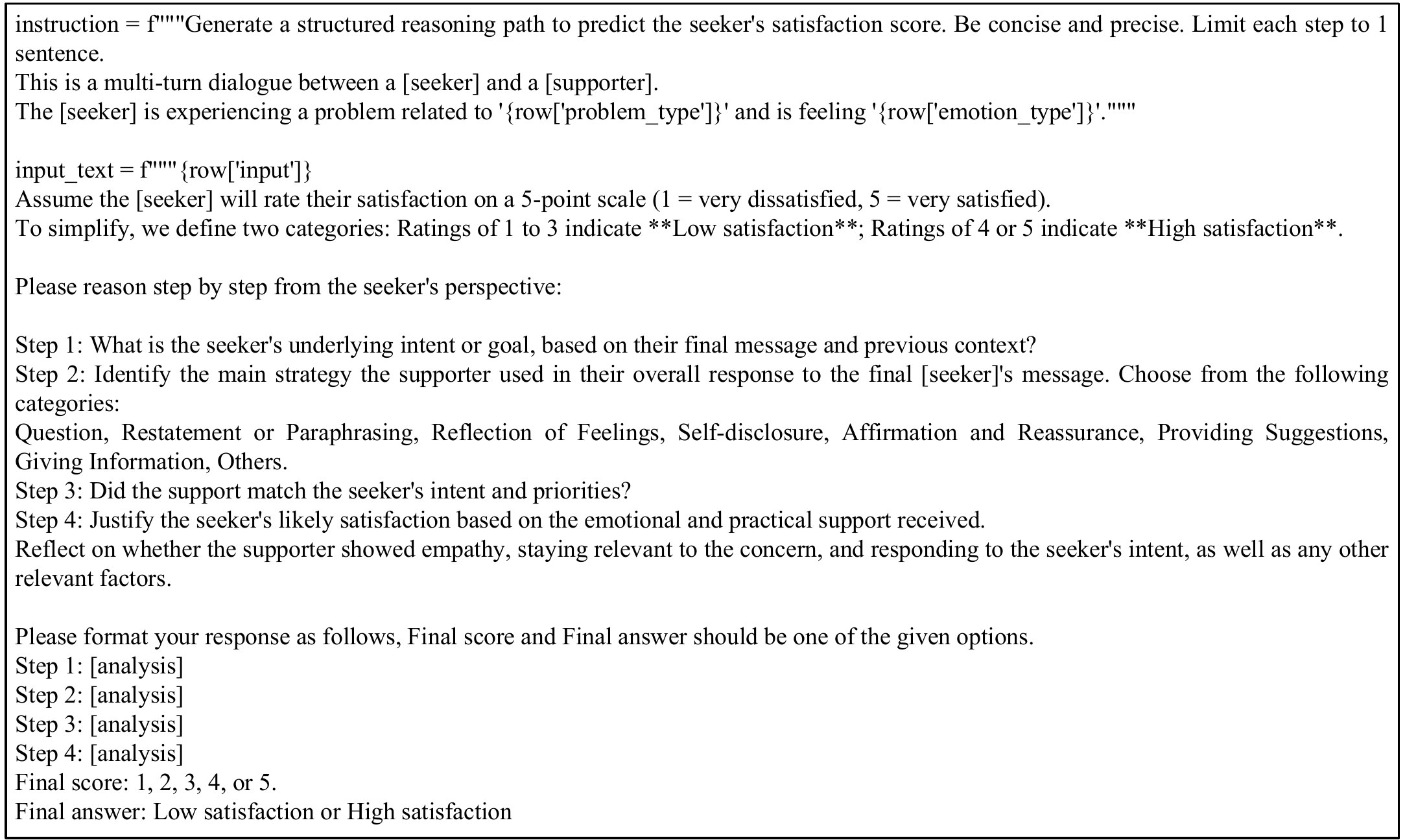}
    \caption{Prompt for User-specific Chain-of-Thought.}
    \label{fig:prompt_cot}
\end{figure*}

\begin{figure*}
    \centering
\includegraphics[width=0.93\textwidth]{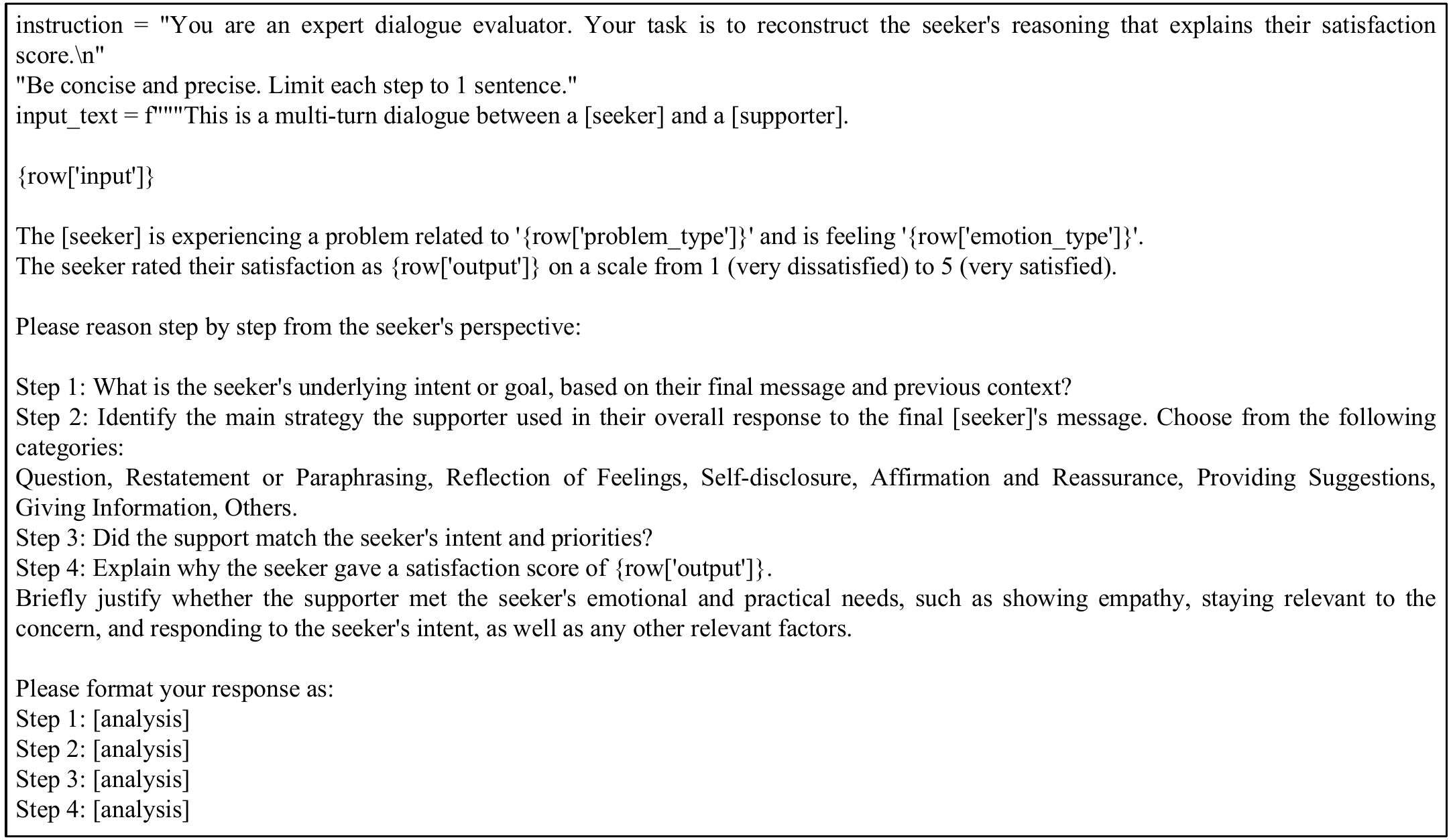}
    \caption{Prompt for User-specific Chains-of-Personalized-Reasoning synthesis.}
    \label{fig:prompt_coper}
\end{figure*}

The prompts used for base input, User-specific Chain-of-Thought, and for synthesizing User-specific Chains-of-Personalized-Reasoning are illustrated in Figure~\ref{fig:base_prompt}, Figure~\ref{fig:prompt_cot}, and Figure~\ref{fig:prompt_coper} respectively.


\section{Analysis of Support Strategies and Feedback Scores}
\label{sec:feedback_dis}
We classify strategies into \textbf{Cognition}-oriented (Question, Restatement or Paraphrasing, Providing Suggestions, Information) and \textbf{Emotion}-oriented (Reflection of Feelings, Self-disclosure, Affirmation and Reassurance), and summarized each strategy's counts and proportions of low (1–3) and high (4–5) feedback scores in Table~\ref{tab:feedback_compare}.

We can see that for both majority and minority groups, emotion-oriented strategies are more likely to receive high feedback (0.94 > 0.91; 0.44 > 0.33), with the effect stronger in the minority group.

\begin{table}[h]
\begin{tabular}{@{}llrr@{}}
\toprule
Group                     & Strategy  & \multicolumn{1}{c}{Low} & \multicolumn{1}{c}{High} \\ \midrule
\multirow{2}{*}{Majority} & Cognition & 302 (0.09)              & 3020 (0.91)              \\
                          & Emotion   & 120 (0.06)              & 1826 (0.94)              \\
\multirow{2}{*}{Minority} & Cognition & 542 (0.67)              & 264 (0.33)               \\
                          & Emotion   & 240 (0.56)              & 192 (0.44)               \\ \bottomrule
\end{tabular}
\caption{Counts and proportion of groups by support strategy and satisfaction score}
\label{tab:feedback_compare}
\end{table}

\begin{figure}[h]
    \centering
\includegraphics[width=\linewidth]{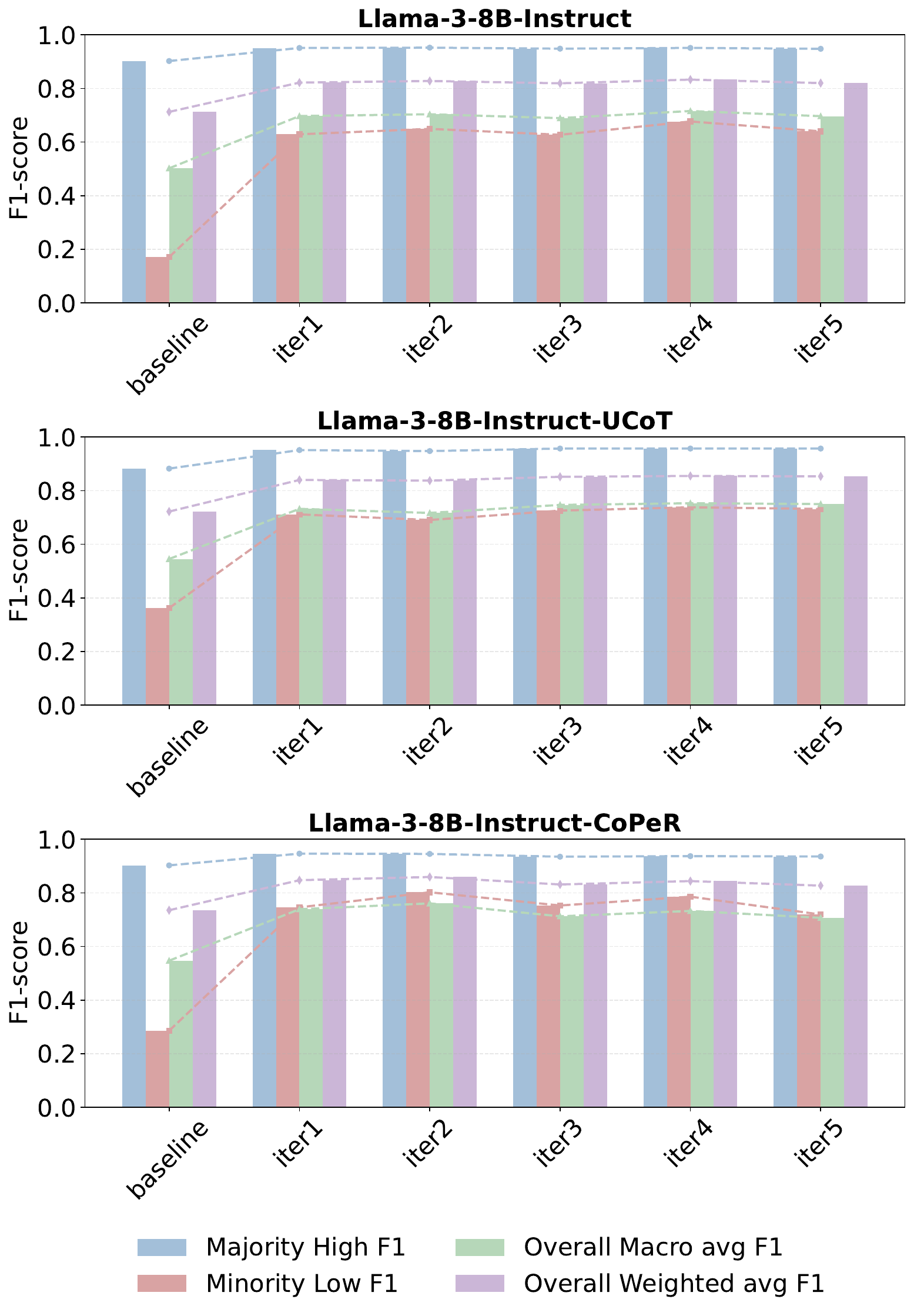}
    \caption{Results on the EM iterations during the Majority-Minority Preference-Aware Clustering stage.}
    \label{fig:em_compare}
\end{figure}

\section{Evaluation of Synthesized Rationales}
\label{sec:evaluation}



\begin{figure*}
    \centering
\includegraphics[width=0.93\textwidth]{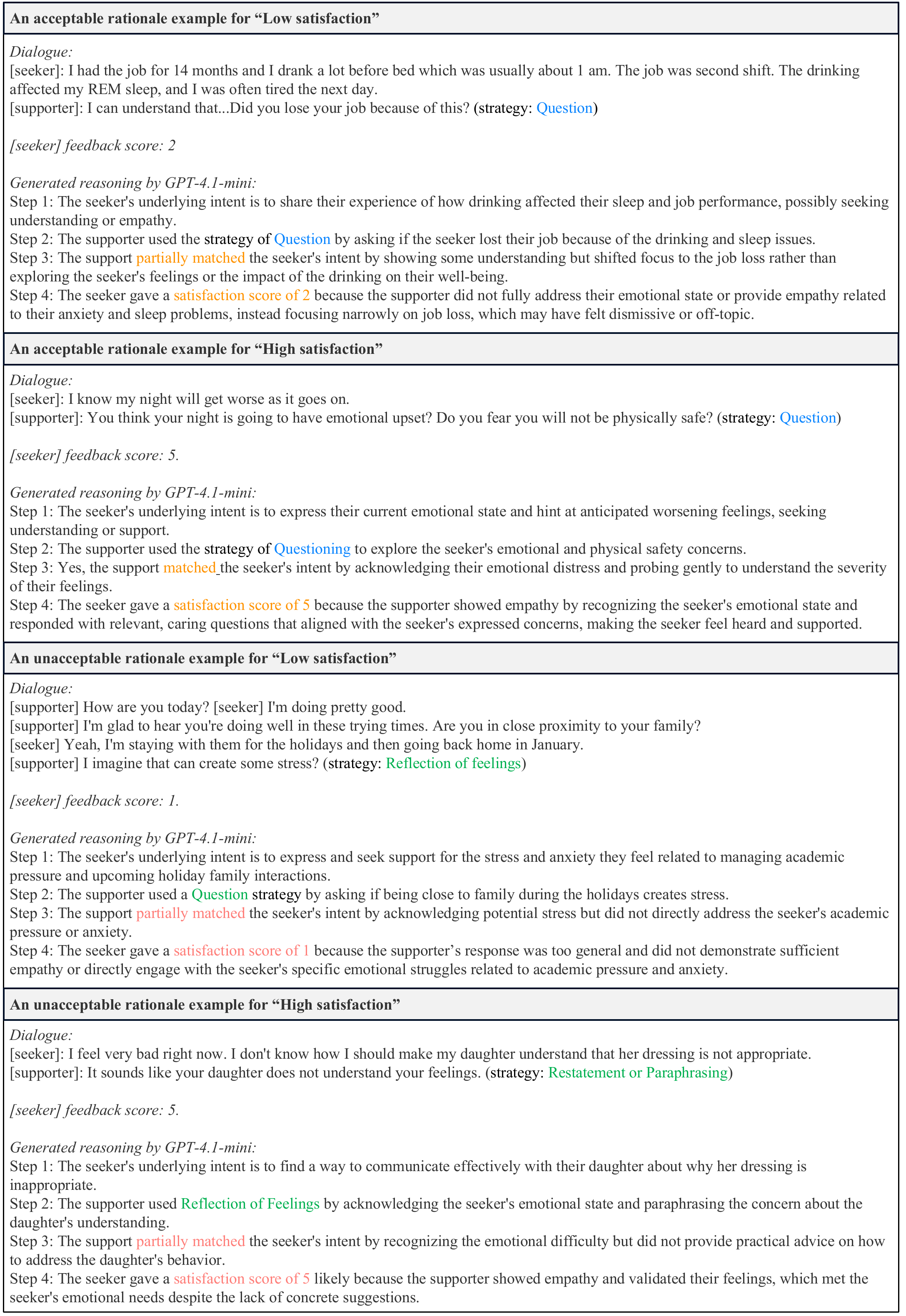}
    \caption{Case Studies.}
    \label{fig:cases}
\end{figure*}

Figure~\ref{fig:cases} presents four case studies: \textit{Low satisfaction} and \textit{High satisfaction} examples under both correct and incorrect supporter-strategy and logical accuracies. 

\section{Results on EM Iterations}
\label{sec:em_iterations}

Figure~\ref{fig:em_compare} presents the detailed results and trends across EM iterations during the Majority-Minority Preference-Aware Clustering stage.



\begin{table*}[h]
  \centering
  \scalebox{0.83}{
  \begin{tabular}{@{}l ll
                  rrrr
                  rrrr
                  rrrr@{}}
    \toprule
    \multirow{2}{*}{Iteration} & \multirow{2}{*}{Users$_{\text{minor}}$} & \multirow{2}{*}{Users$_{\text{major}}$}
    & \multicolumn{4}{c}{Minority}
    & \multicolumn{4}{c}{Majority}
    & \multicolumn{4}{c}{Minority+Majority} \\
    \cmidrule(lr){4-7}\cmidrule(lr){8-11}\cmidrule(lr){12-15}
     & & & $F^{\text{low}}_1$ & $F^{\text{high}}_1$ & $F^{\text{w}}_1$ & $F^{\text{m}}_1$
     & $F^{\text{low}}_1$ & $F^{\text{high}}_1$ & $F^{\text{w}}_1$ & $F^{\text{m}}_1$
     & $F^{\text{low}}_1$ & $F^{\text{high}}_1$ & $F^{\text{w}}_1$ & $F^{\text{m}}_1$ \\
    \midrule

    0  & 10  & 10 & 0.29 & 0.53 & 0.37 & 0.41 & 0.15 & 0.91 & 0.85 & 0.53 & 0.22 & 0.85 & 0.73 & 0.54 \\
    \rowcolor{paleGrayGreen} 
    1  & 10 & 10 & 0.73 & 0.46 & 0.63 & 0.60 & 0.06 & 0.94 & 0.87 & 0.50 & 0.56 & 0.90 & 0.84 & 0.73 \\
    2  & 10 & 10 & 0.72 & 0.40 & 0.61 & 0.56 & 0.09 & 0.94 & 0.88 & 0.52 & 0.56 & 0.90 & 0.84 & 0.73 \\
    3  &  9 & 11 & 0.72 & 0.37 & 0.60 & 0.54 & 0.11 & 0.94 & 0.87 & 0.52 & 0.55 & 0.91 & 0.84 & 0.73 \\
    4  &  8 & 12 & 0.70 & 0.26 & 0.56 & 0.48 & 0.08 & 0.94 & 0.85 & 0.51 & 0.50 & 0.90 & 0.83 & 0.70 \\
    5  &  9 & 11 & 0.71 & 0.30 & 0.57 & 0.51 & 0.03 & 0.94 & 0.86 & 0.48 & 0.52 & 0.90 & 0.83 & 0.71 \\
    \midrule

    0  &  20  &  20  & 0.29 & 0.53 & 0.37 & 0.41 & 0.11 & 0.91 & 0.85 & 0.51 & 0.20 & 0.86 & 0.73 & 0.53 \\
    1  & 18 & 22 & 0.75 & 0.46 & 0.66 & 0.60 & 0.09 & 0.95 & 0.88 & 0.52 & 0.56 & 0.91 & 0.85 & 0.74 \\
    \rowcolor{paleGrayGreen} 
    2  & 17 & 23 & 0.80 & 0.46 & 0.70 & 0.63 & 0.08 & 0.95 & 0.87 & 0.51 & 0.60 & 0.92 & 0.86 & 0.76 \\
    3  & 16 & 24 & 0.75 & 0.31 & 0.62 & 0.53 & 0.00 & 0.93 & 0.85 & 0.47 & 0.52 & 0.90 & 0.83 & 0.71 \\
    4  & 15 & 25 & 0.79 & 0.38 & 0.67 & 0.58 & 0.09 & 0.94 & 0.86 & 0.52 & 0.55 & 0.91 & 0.84 & 0.73 \\
    5  & 17 & 23 & 0.72 & 0.36 & 0.60 & 0.54 & 0.07 & 0.94 & 0.86 & 0.50 & 0.51 & 0.90 & 0.83 & 0.71 \\
    \midrule

    0  & 30   & 30   & 0.29 & 0.53 & 0.37 & 0.41 & 0.11 & 0.91 & 0.85 & 0.51 & 0.20 & 0.86 & 0.73 & 0.53 \\
    \rowcolor{paleGrayGreen} 
    1  & 27 & 33 & 0.73 & 0.40 & 0.63 & 0.57 & 0.12 & 0.95 & 0.88 & 0.53 & 0.56 & 0.91 & 0.85 & 0.74 \\
    2  & 23 & 37 & 0.77 & 0.42 & 0.67 & 0.60 & 0.07 & 0.94 & 0.85 & 0.50 & 0.53 & 0.91 & 0.84 & 0.72 \\
    3  & 25 & 35 & 0.73 & 0.22 & 0.58 & 0.47 & 0.03 & 0.94 & 0.86 & 0.48 & 0.53 & 0.90 & 0.83 & 0.71 \\
    4  & 25 & 35 & 0.73 & 0.29 & 0.60 & 0.51 & 0.03 & 0.94 & 0.86 & 0.49 & 0.54 & 0.91 & 0.84 & 0.72 \\
    5  & 24 & 36 & 0.77 & 0.47 & 0.68 & 0.62 & 0.05 & 0.94 & 0.86 & 0.50 & 0.55 & 0.91 & 0.85 & 0.73 \\
    \midrule

    0  & 40     & 40     & 0.29 & 0.53 & 0.37 & 0.41 & 0.15 & 0.91 & 0.85 & 0.53 & 0.22 & 0.85 & 0.73 & 0.54 \\
    \rowcolor{paleGrayGreen} 
    1  & 34   & 46   & 0.78 & 0.39 & 0.67 & 0.59 & 0.10 & 0.94 & 0.87 & 0.52 & 0.59 & 0.91 & 0.85 & 0.75 \\
    2  & 31   & 49   & 0.78 & 0.46 & 0.69 & 0.62 & 0.00 & 0.94 & 0.85 & 0.47 & 0.53 & 0.91 & 0.84 & 0.72 \\
    3  & 31   & 49   & 0.74 & 0.35 & 0.63 & 0.55 & 0.03 & 0.94 & 0.85 & 0.48 & 0.51 & 0.90 & 0.83 & 0.71 \\
    4  & 33   & 47   & 0.76 & 0.42 & 0.66 & 0.59 & 0.03 & 0.94 & 0.86 & 0.48 & 0.54 & 0.91 & 0.84 & 0.72 \\
    5  & 27   & 53   & 0.83 & 0.47 & 0.73 & 0.65 & 0.02 & 0.93 & 0.83 & 0.48 & 0.53 & 0.91 & 0.84 & 0.72 \\
    \midrule

    0  & 50     &  50    & 0.29 & 0.53 & 0.37 & 0.41 & 0.11 & 0.91 & 0.85 & 0.51 & 0.20 & 0.86 & 0.73 & 0.53 \\
    1  & 40   & 60   & 0.71 & 0.35 & 0.59 & 0.53 & 0.05 & 0.94 & 0.85 & 0.49 & 0.50 & 0.90 & 0.82 & 0.70 \\
    2  & 43   & 57   & 0.73 & 0.47 & 0.64 & 0.60 & 0.08 & 0.94 & 0.87 & 0.51 & 0.55 & 0.91 & 0.84 & 0.73 \\
    3  & 37   & 63   & 0.83 & 0.47 & 0.73 & 0.65 & 0.11 & 0.93 & 0.85 & 0.52 & 0.58 & 0.91 & 0.85 & 0.75 \\
    \rowcolor{paleGrayGreen}
    4  & 42   & 58   & 0.77 & 0.36 & 0.63 & 0.57 & 0.06 & 0.94 & 0.87 & 0.50 & 0.59 & 0.91 & 0.85 & 0.75 \\
    5  & 36   & 64   & 0.77 & 0.38 & 0.64 & 0.58 & 0.13 & 0.93 & 0.85 & 0.53 & 0.55 & 0.91 & 0.84 & 0.73 \\
    \bottomrule
      \end{tabular}}
  \caption{Results of the proposed \textbf{Llama-3-8B-CoPeR +  M\textsuperscript{2}PC}, when initializing majority and minority users as 10, 20, 30, 40, and 50, respectively.}
  \label{tab: users}
\end{table*}

\section{Results on Different User Clusters}
\label{sec:ablation}
For training, we randomly selected from the majority group the same number of conversations as in the minority group, with each conversation representing a user. These users are evenly and randomly divided into 20 clusters per group. During M²PC training, each cluster (a batch of users) is reassigned to majority or minority groups based on perplexity in an unsupervised manner; therefore, no universally accepted partition rule is required. We believe this approach may help capture finer subpopulation preferences within each majority or minority group, as the routing perplexity is computed from the average perplexity of all users in each cluster/subpopulation.
Table~\ref{tab: users} presents the results of the proposed \textbf{Llama-3-8B-CoPeR + M\textsuperscript{2}PC} model when the numbers of majority and minority users are initialized to 10, 20, 30, 40, and 50, respectively. The best performance is obtained when both groups are initialized to 20, with which after the second iteration, the algorithm yields 23 distinct majority groups and 17 distinct minority groups, indicating that M\textsuperscript{2}PC adaptively clusters users with similar preferences.

\end{document}